%% file: emnlp2020.tex
\pdfoutput=1
\documentclass[11pt]{article}

\usepackage{emnlp2021}

\usepackage{times}
\usepackage{latexsym}
\usepackage{booktabs}

\usepackage[T1]{fontenc}
\usepackage[utf8]{inputenc}

\usepackage{microtype}
\usepackage{natbib}
\usepackage{caption}

\include{preamble}
\usepackage{setspace}
\usepackage[a4paper]{geometry}
\usepackage{titlesec}
\usepackage{amssymb}
\usepackage{mathtools}
\usepackage{lipsum}

\usepackage{amsmath}

\usepackage{multirow}      
\usepackage{multicol}
\usepackage{pgfplots}
\newcommand\Tstrut{\rule{0pt}{2.6ex}}       % "top" strut
\newcommand\Bstrut{\rule[-0.9ex]{0pt}{0pt}} % "bottom" strut
\newcommand{\TBstrut}{\Tstrut\Bstrut} % top&bottom struts
\PassOptionsToPackage{hyphens}{url} % Faeze added for url small font

\usepackage[normalem]{ulem}
\usepackage{comment}
\usepackage{blindtext, subfig} 
\usepackage[linecolor=orange]{todonotes}
\usepackage{hhline}
\usepackage{booktabs} % To thicken table lines
\usepackage{enumitem}% to reduce space between items.

\newcommand{\rebart}{\textsc{Re}-\textsc{Bart}\xspace}

\title{\emph{Is Everything in Order?} A Simple Way to Order Sentences}

\author{Somnath Basu Roy Chowdhury$^{1}$\thanks{ \hspace{0.08cm} Authors contributed equally.} \qquad Faeze Brahman$^{2*}$ \qquad Snigdha Chaturvedi$^{1}$ \\ $^{1}$UNC Chapel Hill, $^{2}$University of California, Santa Cruz \\ \texttt{\{somnath, snigdha\}@cs.unc.edu \qquad \texttt{fbrahman@ucsc.edu}}}

\date{}

\begin{document}
\maketitle

\begin{abstract}

The task of organizing a shuffled set of sentences into a coherent text has been used to evaluate a machine's understanding of causal and temporal relations. We formulate the \textit{sentence ordering} task as a conditional text-to-marker generation problem. We present \textbf{Re}order-\textbf{BART} (\rebart) that leverages a pre-trained Transformer-based model to identify a coherent order for a given set of shuffled sentences. The model takes a set of shuffled sentences with sentence-specific markers as input and generates a sequence of position markers of the sentences in the ordered text. \rebart achieves the state-of-the-art performance across 7 datasets in Perfect Match Ratio (PMR) and Kendall's tau ($\tau$). We perform evaluations in a zero-shot setting, showcasing that our model is able to generalize well across other datasets. We additionally perform several experiments to understand the functioning and limitations of our framework.

%%% older version (May 14)
% The task of organizing a shuffled set of sentences into a coherent text has been used to evaluate a machine's understanding of causal and temporal relations.  We present \textbf{Re}order-\textbf{BART} (\rebart), a \emph{sentence ordering} method 
% that leverages a pre-trained Transformer-based model to identify a coherent order for a given set of shuffled sentences. We formulate the task as a conditional text-to-marker generation problem where the input is a set of shuffled sentences with sentence-specific markers and the output is a sequence of position markers of the sentences in the ordered text.\sbrc{this sentence is too long} \rebart achieves the state-of-the-art performance across 7 datasets in Perfect Match Ratio (PMR) and Kendall's tau ($\tau$). We perform evaluations in a zero-shot setting, showcasing that our model is able to generalize well across other datasets. We additionally perform several experiments to understand the functioning and limitations of our framework.

\end{abstract}

\section{Introduction}

% \fbb{remember to add significant test for our results. }

 % Constructing coherent text requires an understanding of entities, events, and their relationships.  Automatically understanding the relationships among nearby sentences in a multi-sentence text, like temporal and causal relations, is a longstanding problem in NLP.
 Constructing coherent text requires an understanding of entities, events, and their relationships.  Automatically understanding such relationships among nearby sentences in a multi-sentence text has been a longstanding challenge in NLP.
 
 %\fbb{Agree. How about swapping the order of the two? I am also fine with removing the second sentence and merge the first sent with next para}\sbrc{thanks, done!}
 %should respect the logical order of events so as to make comprehension easier.
%\SC{You are focusing too much on events, even though you don't do anything specific about events. Make it broader. Say something like it needs understanding of entities, events, and relationships between them. Read about the centering theory, and check out this paper: https://www.aclweb.org/anthology/J08-1001.pdf}. %which helps in better comprehension of the text.

Sentence ordering task was proposed to test the ability of automatic models to reconstruct a coherent  text given a set of shuffled sentences~\cite{barzilay2008modeling}. 
Coherence modeling has wide applications in natural language generation like extraction-based multi-document summarization \cite{barzilay2002inferring, galanis2012extractive, nallapati2017summarunner}, retrieval dependent QA \cite{yu2018qanet, liu2017stochastic} and concept-to-text generation \cite{schwartz2017effect}.

Earlier studies on coherence modeling and sentence ordering focused on exploiting different categories of features like coreference clues~\cite{elsner-charniak-2008-coreference}, entity grids~\cite{LapataB05, barzilay2008modeling}, named-entity categories~\cite{elsner2011extending}, and syntactic features~\cite{louis-nenkova-2012-coherence} among others. With the advent of deep learning, researchers leveraged distributed sentence representations learned through recurrent neural networks~\cite{li-hovy-2014-model}. Recent works adopted ranking-based algorithms to solve the task~\cite{chen2016neural, kumar2020deep, prabhumoye2020topological}. 
%\sbrc{this paragraph is exactly same as the related works section, why do we have this?} \fbb{when I wrote this para, this was not part of the related work. Also the intro should have some sense of related work. If you use the same phrasing in related work, pls paraphrase that.}

\begin{figure}[t!]
	\centering
	\includegraphics[width=0.5\textwidth]{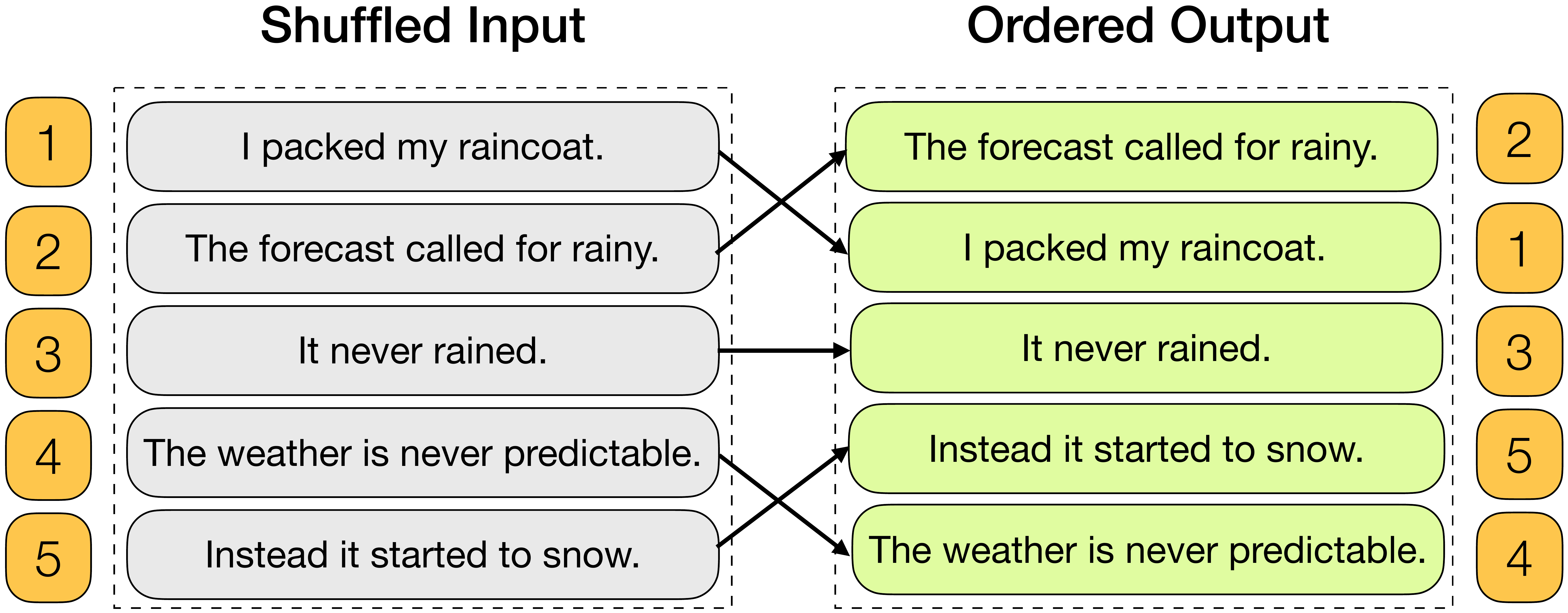}
	\caption{ An example of the sentence ordering task. The goal is to reorder a set of shuffled sentences (left) into a coherent sequence of text (right).} %\SC{Not a big deal but if possible use a less fluorescent green. Also, it should be "The forecast called for rain".}
	\label{tab:examples}
\end{figure}

In this paper, we present \textsc{Re}-\textsc{Bart} (for \textbf{Re}order-\textbf{BART}) to solve the sentence ordering as a conditional text-to-marker generation where the input is a shuffled set of sentences and output is a sequence of position markers for the coherent sentence order. %\fbb{should we refer to fig 2 here?}.\sbrc{personally: I feel it's a bit premature here, no strong feelings though}

\begin{figure*}[t!]
	\centering
	\includegraphics[width=0.95\textwidth]{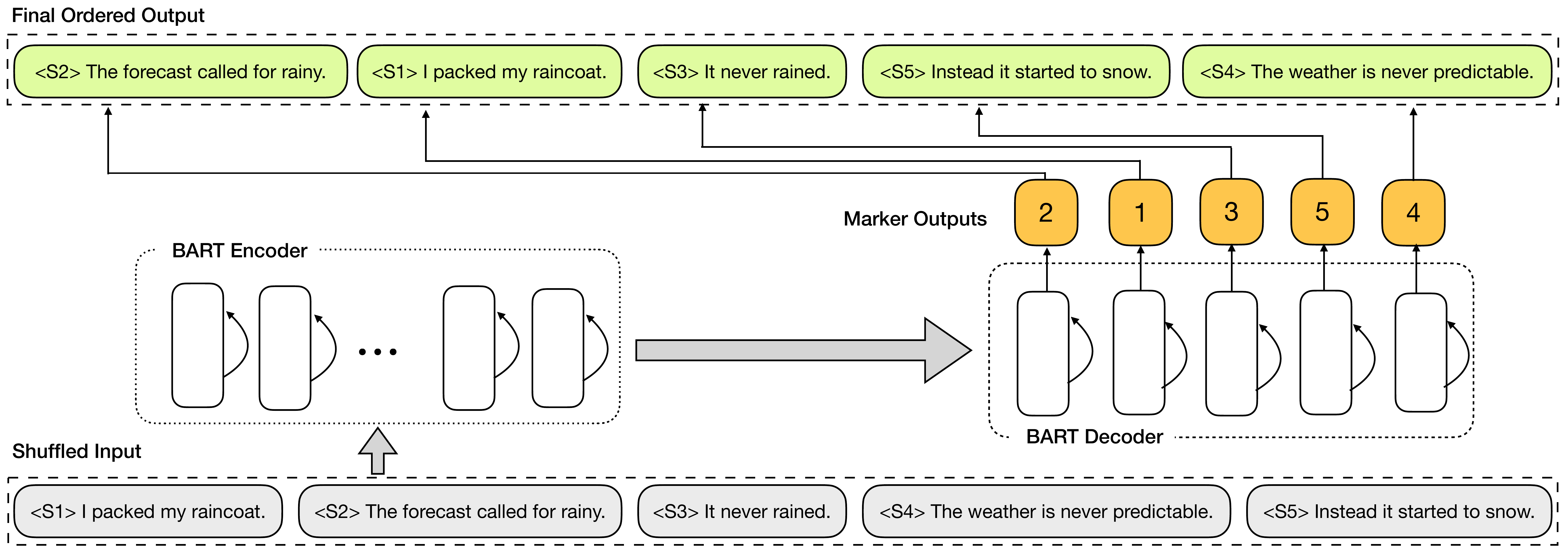}
	\caption{Proposed \rebart framework. Given a shuffled set of input sentences, {\rebart} generates Marker Outputs (position markers of sentences) which is then used to reconstruct the ordered output. }
	\label{fig:setup}
\end{figure*}

Sentence ordering can be viewed as a task of reconstructing the correct text from a noisy input. For this reason we use BART \cite{lewis2019bart} as the underlying generation module for {\rebart}. %\fbb{Can we say "Since sentence ordering can be viewed as ..., we use BART...?" The reason is because the first sentence as it is is very repetitive and the reader saw that by now several times. It's annoying as a single sentence. } \sbrc{this is different from conditional text generation setup, and is not repeated elsewhere. Also not good to start a para with "since"}
BART is pre-trained as a denoising autoencoder where one of the objective involves generating the coherent text from corrupted input sequences. %\fbb{This sounds like sentence permutation is the only objective of BART.}%\SC{It is still not clear why you are telling me this fact. why is this information about BART is relevant for understanding your approach? It's like telling me BART is pre-trained for X hours on blah blah blah machine. As a reader, why do I care about it? General rule: instead of only providing a fact (BART is pretrained on blah) \textbf{explicitly} state how that fact is relevant here/what conclusions can you draw. Since you are the author, the burden of explanation is on you. Do not rely on the reader to filter out the relevant information from a sentence and connect it to your story. In this case, either remove this sentence (fact) or say why  corrupted input sequences is important (maybe it is similar to your problem?)}. 
Prior works encode sentences individually or in a pairwise manner and then compute the position of a sentence in the paragraph. We instead encode the entire shuffled sequence at once, which results in learning better 
token representations with respect to the entire input context.  This helps the model in capturing interactions among all sentences and identifying the relative order among them. % \SC{What does it mean to encode x as input? What does it mean to identify the relative order in something shuffled?}

% \SC{The figure says "Marker Output" which is not explanatory enough. You could use the caption to describe the input and output: Given this as input the model generates blah }

Our simple framework outperforms previous state-of-the-art by a large margin for all benchmark datasets. % in Perfect Match Ratio (PMR) and Kendall's tau ($\tau$) metric.
Specifically, we achieve 11.3\%-36.2\% relative improvements in Perfect Match Ratio (PMR) and 3.6\%-13.4\% relative improvements in Kendall's tau ($\tau$) across all benchmarks. Our main contributions are: %\sout{fourfold}: this mean 4 times
\begin{itemize}[noitemsep,topsep=0.3pt, leftmargin=*]
    \setlength\itemsep{0.3em}
    \item We formulate the \textit{sentence ordering} as a conditional text generation problem and present a simple method to solve it. %We propose a simple framework to solve the \textit{sentence ordering task} as conditional text generation, which is easy to implement and less prone to neural text degeneration.
    \item We empirically show that our model significantly outperforms existing approaches by a large margin and achieves state-of-the-art performances across all benchmark datasets.
    % establish new state-of-the-art, outperforming existing approaches by large margins.
    \item We conduct zero-shot evaluations showing our model trained on Movie Plots outperforms the previous in-domain trained state-of-the-art.
    \item We present a thorough analysis to evaluate sensitivity of our model to different input properties.
\end{itemize}

\section{Related Work}
\label{related-w}

The problem of sentence ordering can be formulated as finding an order with maximum coherence. Earlier works focused on modeling local coherence using linguistic features \cite{LapataB05, barzilay2008modeling, elsner2011extending, guinaudeau2013graph}.

% \SC{this introductory sentence is good. }
%\noindent \textbf{Pointer networks \& pairwise models}:  
A line of work have leveraged neural networks to encode sentences and retrieve the final order using pointer network \cite{vinyals2015pointer} by comparing them in a pairwise manner~\cite{gong2016end, logeswaran2018sentence, cui2018deep, yin2019graph, yin2020enhancing}. HAN \cite{wang2019hierarchical} and TGCM \cite{oh2019topic} used an attention based pointer network for decoding. B-TSort \cite{prabhumoye2020topological}  uses topological sorting to retrieve the final order from sentence pairs. \citet{zhu2021neural} encode sentence-level relationships as constraint graphs to enrich sentence representations. 
The state-of-the-art approach \cite{cui2020bert} introduced a novel pointer decoder with a deep relational module. 

%\noindent \textbf{Ranking-based frameworks}:
Other works considered reframing the task as a ranking problem.
\citet{chen2016neural} proposed a model which relies on a ranking framework to retrieve the order of sentence pairs. \citeauthor{kumar2020deep} (\citeyear{kumar2020deep}) utilized a BERT \cite{devlin2018bert} encoder to generate scores for each sentence which were used to sort them into the correct order. 

Different from these approaches, we formulate sentence ordering as a conditional text generation task. We use a sequence-to-sequence model in our framework where the decoder encapsulates the functioning of a pointer network while generating output sentence positions. Our code is available at: \href{https://github.com/fabrahman/ReBART}{https://github.com/fabrahman/ReBART}.

\section{\rebart}

%We now start by formulating the sentence ordering task and then present our \rebart model.
%\SC{If you need space you can get rid of the sub-headings in this section and this first introductory sentence}
%\subsection{Task Formulation}
Given a sequence of shuffled sentences $S' = \{s'_1, s'_2, \ldots, s'_{N_S}\}$, where $s'_i$ denotes the $i^{th}$ sentence and $N_S$ denotes the number of input sentences, the task is to generate the ordered output sequence $S^* = \{s_1, s_2, \ldots, s_{N_S}\}$.

We solve the sentence ordering task using a text-to-marker framework shown in Figure~\ref{fig:setup}. % which is shown in Figure~\ref{fig:setup}. 
Specifically, taking a shuffled sequence of sentences ($S'$) as input, we generate a sequence of {position markers} $Y = \{y_1, y_2, \ldots, y_{N_S} \}$ as output, where $y_i$ denotes the position of $i^{th}$ sentence of the corresponding ordered sequence ($s_i$) 
%\SC{Is this correct? How is si related to S' and $\hat{S}$}\fbb{si is each sentence in the original ordered input which we seek to find its position in the shuffled sequence} 
in the shuffled input. %\fbb{Is the last sentence not clear?}
The ordered output sequence can then be reconstructed as $\hat{S} = \{S'_{y_1}, S'_{y_2}, \ldots, S'_{y_{N_S}}\}$. % Figure \ref{fig:setup} shows an example of our setup. %\SC{new para} 

Our goal is to train a probabilistic model $P_{\theta}(\mathbf{Y}|\mathbf{S'})$ %which takes as input a shuffled set\SC{You have said this too many times by now} of sentences $S'$ and generates a sequence of position markers $Y$, 
% with the following optimization objective:
by optimizing:
\begin{equation}
\max_{\theta} \log P_{\theta}(\mathbf{Y}|\mathbf{S'})    
\end{equation}

The functioning of \rebart model is shown in Figure~\ref{fig:setup}. \rebart consists of a sequence-to-sequence model with an encoder to receive a shuffled set of sentences, and a decoder to generate \textit{position markers} (2, 1, 3 etc.), which is then used to retrieve the final ordered sequence.
We use BART~\cite{lewis2019bart} as the underlying sequence-to-sequence model, since our task can benefit from the \textit{sentence permutation} pre-training objective. Additionally, to provide the model with a %better \SC{either don't use better, or specify better than what?}
supervision signal to generate position markers, we append each sentence in the shuffled input with \textit{sentence markers} (\texttt{\textless S1\textgreater, \textless S2\textgreater},  etc.).\footnote{We experimented with various combinations of sentence markers and position marker, and found out that the text-to-marker framework performs the best.} Sentence markers were added as special tokens to the tokenizer. \rebart learns to attend to the markers while generating the final order $Y$.

The proposed text-to-marker framework has two advantages over an alternate text-to-text framework, where the model directly generates the entire text sequence instead of marker outputs. First, the model performs better as the output space is much smaller.  This also makes it less susceptible to neural text degeneration \cite{holtzman2019curious},  as significantly fewer output tokens are generated. % \SC{you need one more sentence here connecting generating fewer tokens to lesser degradation }
Second, when generating the entire text sequence in the text-to-text framework, we observe that the model often generates text which is not part of the input, rendering the output invalid for the task.

% \SC{Are there any other details to be mentioned about the model or model training. Currently this section looks short}

\section{Experimental Setup}
% We first describe several benchmark sentence ordering datasets. We then explain 
% In this section, we evaluate the performance of {\rebart} on several benchmark sentence ordering datasets. We also conduct a series of experiments to better understand the working of our model and investigate its generalization capability. \SC{Is this correct. This section is only describing the evaluation setup.  }

\begin{table}[t!]
	\begin{center}
		\resizebox{0.5\textwidth}{!}{
        \input{figures/dataset_stat}
		}
	\end{center}
% 	\vspace{-8pt}
	\caption{Dataset statistics.} %\sbrc{should we report numbers like 2K, 8k? to make it more readable}\SC{either way is fine}
	\label{tab:dataset-stats}
\end{table}

\subsection{Datasets}
%\fbb{This section we use "we evaluate our model" three times. How about the first one to be: "we run our experiments on 7 ..."}
%\fbb{We can remove the footnotes on dataset urls and move it to appendix to save space.} \SC{If you provide citations for the datasets in the main text, you don't need the urls}
We run our experiments on {7} publicly available English datasets from two domains: scientific paper abstracts and narratives. 

\noindent \textbf{Paper Abstracts:} We evaluate our model on 4 datasets, obtained from abstracts of scholarly articles. The datasets include abstracts from NeurIPS, AAN, ACL, NSF Research Award, and arXiv~\cite{LogeswaranLR18, gong2016end, chen2016neural}. % \footnote{https://arxiv.org/}

\noindent \textbf{Narratives:} We evaluate our model on 3 datasets in the narrative domain. %\SC{not a big deal but earlier you said datasets are obtained from abstracts... not you say SENTENCES are from a narrative make me wonder why you have different wordings, are sentences not continuous in the original narrative etc. Just say "datasets based on narratives"}. %story/movie \SC{why use /? movie is also a story? Also the term "movie narrative" or "story narrative" does not make sense.} 
ROCStories~\cite{huang2016visual} contains five-sentence long stories about everyday events. % a commonsense story dataset, which is randomly split into training/test/validation sets in a 80:10:10 ratio.
SIND\footnote{http://visionandlanguage.net/VIST/dataset.html}~\cite{huang2016visual} is a visual storytelling dataset. Wikipedia Movie Plots \footnote{www.kaggle.com/jrobischon/wikipedia-movie-plots} contains plot description of movies from Wikipedia.\footnote{Movie plots contains instances with long paragraphs, we consider the first 20 sentences in every instance.}
% \SC{You talk about train-test-split for RocStories but not for others? Why is Rocstories special?} \sbrc{yeah RocStories is special as it doesn't have a pre-defined train/test split} \SC{Talk about the train-test splits of rocstories in the last para where you talk about train, test sets for other datasets. } 

We randomly split ROCStories into train/test/validation in a 80:10:10 ratio. For the other datasets, we use the same train, test and validation sets as previous works. Dataset statistics are reported in Table~\ref{tab:dataset-stats}.

\subsection{Implementation Details}
We use Huggingface library~\cite{Wolf2019HuggingFacesTS} for our experiments. During inference we decode the output positions in a greedy manner by choosing the logit with the highest probability. The hyper-parameters used for each dataset are provided in {Table 1} in the Appendix. The experiments are conducted  in PyTorch framework using Quadro RTX 6000 GPU. The hyper-parameters for each dataset are provided in Table~\ref{tab:hyperparams}.

\begin{table}[t!]
	\begin{center}
% 	\footnotesize
		\resizebox{0.48\textwidth}{!}{
        \input{figures/hyperparams}
		}
	\end{center}
	\caption{Hyperparameter settings on each dataset.}
	\label{tab:hyperparams}
\end{table}
% \footnote{https://huggingface.co/}

\begin{table*}[h!]
	\begin{center}
		\resizebox{\textwidth}{!}{
\input{figures/results}
		}
	\end{center}
\end{table*}
\begin{table*}[h!]
	\begin{center}
% 	\small
	\vspace{-15pt}
		\resizebox{0.9\textwidth}{!}{
\input{figures/results-story}
		}
% 		\vspace{-5pt}
		\caption{Performance on abstracts (top) and narratives (bottom) datasets. The best and second-best scores are in bold and underlined. {\rebart} achieves the state-of-the-art performance in PMR and $\tau$ for all datasets.}
		\label{results}
	\end{center}
\end{table*}

\subsection{Evaluation Metrics}
\label{sec:metrics}
Following previous works \cite{cui2020bert, kumar2020deep, wang2019hierarchical}, we use the following metrics for evaluating our approach: %\SC{"on sentence ordering task" is unnecessary. Unless you have multiple tasks, by now it should be obvious to the reader what the task is. If it isn't you need to revise your paper :)}:

    \noindent \textbf{Accuracy} (Acc): This is the fraction of output sentence positions predicted correctly averaged over all test instances. It is defined as: %In our setup \SC{"In our setup" is unnecessary because there is only one setup you have been talking about and the definition in the previous sentence is very specific to your setup}
    % \fbb{Is there any reason you have numbered equation only for acc?}
    \begin{equation*}
        \mathrm{Acc} = \mathop{\mathbb{E}}\limits_{S' \sim \mathcal{D}} \bigg[\frac{1}{N_S} \sum\limits_{i=1}^{N_S} \mathbb{I}(S'_{y_i} = s_i)\bigg]
    \end{equation*}
    where $S'$ is a shuffled input from the dataset $\mathcal{D}$, $s_i$ is the $i^{th}$ sentence in the ordered sequence, $y_i$ is the predicted sentence marker at position $i$ and $N_S$ is the number of sentences in the input. %\SC{also define yi for completeness}.

    \noindent \textbf{Perfect Match Ratio} (PMR): PMR measures the fraction of sentence orders exactly matching with the correct order across all input instances:
    
    % \mathbb{E}_{S' \sim \mathcal{D}}
    $$\mathrm{PMR} = \frac{1}{N}\sum\limits_{j=1}^{N}\bigg[\mathbb{I}({Y_j} = Y_j^*)\bigg]$$
    % \noindent where $Y$ and $Y^*$ are the predicted and gold position marker sequences, respectively, and $N$ is the number of instances in the dataset.    
    \noindent where $Y_j$ and $Y_j^*$ are the predicted and gold position marker sequences, respectively, and $N$ is the number of instances in the dataset.

    \noindent\textbf{Kendall's Tau} ($\tau$): $\tau$ is a metric to evaluate the correlation between two sequences:
    $$\tau = 1 - \frac{2 \;(\#\; \mathrm{inversions})}{\big(\frac{n}{2}\big)}$$
    In our setup, we evaluate $\tau$ between the predicted position marker sequence $Y$  and gold position marker sequence $Y^*$. 
    % \SC{If you use Y* for gold why not use S* for gold} \fbb{should we use $\hat{Y}$ for all predictions? If so we need to be careful in replacing all instances of that.} \sbrc{too late, also will mess up section 2}
    A higher score indicates a better performance for all metrics.

\subsection{Baselines}
 We compare \rebart with 11 previous sentence ordering frameworks including the current state-of-the-art BERSON \cite{cui2020bert}. Other baselines include  B-TSort \cite{prabhumoye2020topological}, RankTxNet \cite{kumar2020deep}, TGCM \cite{oh2019topic}, FUDecoder \cite{yin2020enhancing}, SE-Graph \cite{yin2019graph}, HAN \cite{wang2019hierarchical}, ATTOrderNet \cite{cui2018deep}, V-LSTM+PtrNet \cite{logeswaran2018sentence}, LSTM+PtrNet \cite{gong2016end} and Pairwise model \cite{chen2016neural}. Apart from these baselines, we also include a text-to-text variant of our model, where we fine-tune a pre-trained BART model to generate the  text sequences corresponding to sentences instead of their markers. We call this variant BART (fine-tuned).

\section{Results}
In this section, we evaluate the performance of {\rebart} on several benchmark sentence ordering datasets. We also conduct a series of experiments to {better} understand the working of our model and investigate its generalization capability.

Table~\ref{results} reports the experimental results on all benchmark datasets.\footnote{%Results from prior work 
Prior results have been compiled from \cite{cui2020bert}.} %\SC{stuff like this goes in the footnote}. 
\rebart improves over all baselines by a significant margin and achieves the new state-of-the-art results in PMR and $\tau$ metrics for all datasets. In particular, \rebart  improves the previous state-of-the-art performance in PMR metric by a relative margin of 18.8\% on NeurIPS,  22.9\% on AAN, 28.9\% on NSF, 11.3\% on arXiv, 36.2\% on SIND and 20\% on ROCStories. We observe similar relative gains in $\tau$: 4.7\% on NeurIPS, 7.1\% on AAN, 13.4\% on NSF, 3.6\% on arXiv, 10.8\% on SIND and 6.8\% on ROCStories.  %\sbrc{need to verify all the numbers} DONE!

We observe that \rebart's performance on Wikipedia Movie Plots is relatively poor compared to other datasets. This could be because this dataset has relatively longer input sequences (Table~\ref{tab:dataset-stats}), making the task more challenging for the model.

\begin{table}[t!]
	\begin{center}
		\resizebox{0.5\textwidth}{!}{
        \input{figures/text_vs_index}
		}
% 		\vspace{-6pt}
		\caption{ Model performance using text-to-text and text-to-marker frameworks on ROCStories. A significant gain is observed using text-to-marker framework.
		}
		\label{index-vs-text}
	\end{center}
\end{table}

\noindent\textbf{Comparison with text-to-text framework:}
%\SC{Why isn't this subsection appearing as a continuation of main results?} 
% \SC{Why not talk about the following results in this section which is titled comparison with text-to-text framework? I am moving it here: Table ??? also shows that} 
Table~\ref{results} also shows that \rebart outperforms BART (fine-tuned), our text-to-text baseline, for all datasets. BART (fine-tuned) performs reasonably well on the NeurIPS, AAN, SIND and ROCStories datasets where the average number of sentences (Table~\ref{tab:dataset-stats}) is low. It struggles on NSF, arXiv and Movie Plots where input sequences are longer.
Upon manual inspection, we found that BART (fine-tuned) model suffers from neural text degeneration~\cite{holtzman2019curious} and produces output tokens which aren't present in the input.

%\sbrc{this transition isnt smooth, any suggestions?} \fbb{why can't we put the following para in BART vs. T5 subsection?}
We hypothesize that training in our proposed text-to-marker framework yields a performance gain over text-to-text framework, irrespective of the underlying sequence-to-sequence model. 
To verify this hypothesis, we compare two settings of our framework that use BART and T5 as the underlying sequence-to-sequence model. % We perform this experiment on the ROCStories dataset \SC{You are inviting criticism by having a statement like this: why not on other datasets? Either provide a convincing reason or try to blend this information in one of the other sentences, captions, etc}.
 In Table~\ref{index-vs-text}, we observe significant gains for both BART and T5 using our text-to-marker framework. This shows the text-to-marker framework outperforms text-to-text baseline irrespective of the generation module. %\SC{This sentence is talking about two conclusions that one can draw: (1) the marker framework is better irrespective of the s2s model, (2) BART is better irrespective of the framework. Don't mix them. Dedicate different paragraphs to both. }. 

% \sbrc{this paragraph starting feels completely out of place}
%\SC{You provide two reasons here. However, reason 2 is not easy to understand without looking at comparison with text-to-text framework. You should put this para after that comparison.} 
From results in Table~\ref{results}, we observe that our simple framework is effective and outperforms more complex baseline architectures. One explanation behind \rebart's success could be the use of sentence markers. \rebart is able to encapsulate the context in individual sentences (observed in generated attention maps in \S\ref{sec:attention}) and produce markers at the correct output position. Additionally, our text-to-marker framework is better at leveraging causal and temporal cues implicitly captured by BART during pre-training. 

% \SC{"causal and temporal cues" is not supported anywhere. You are claiming more than you can support. Also, not a big deal but you are just saying that the improvement is because of BART (aka you didn't do much). I would start with the more interesting reason (the second one). }

\begin{figure}[t!]
\centering
\vspace{-0.3cm}

\subfloat[][\footnotesize BART embeddings]{
    \includegraphics[width=0.2\textwidth]{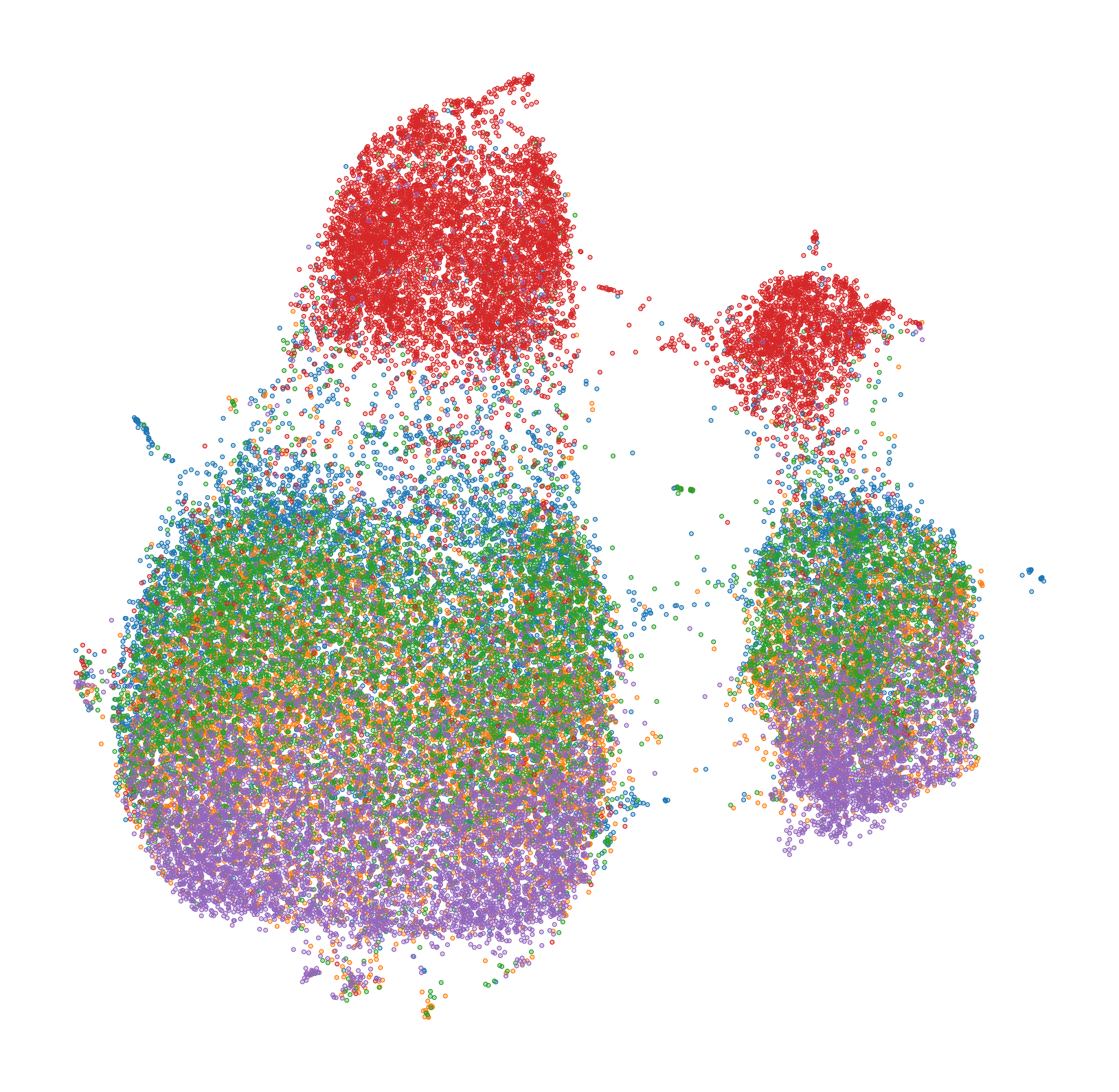}{ % 0.31
    \label{bart-emb}}
    } 
    % \vspace{-15pt}
\subfloat[][\footnotesize T5 embeddings]{ %
    \includegraphics[width=0.2\textwidth]{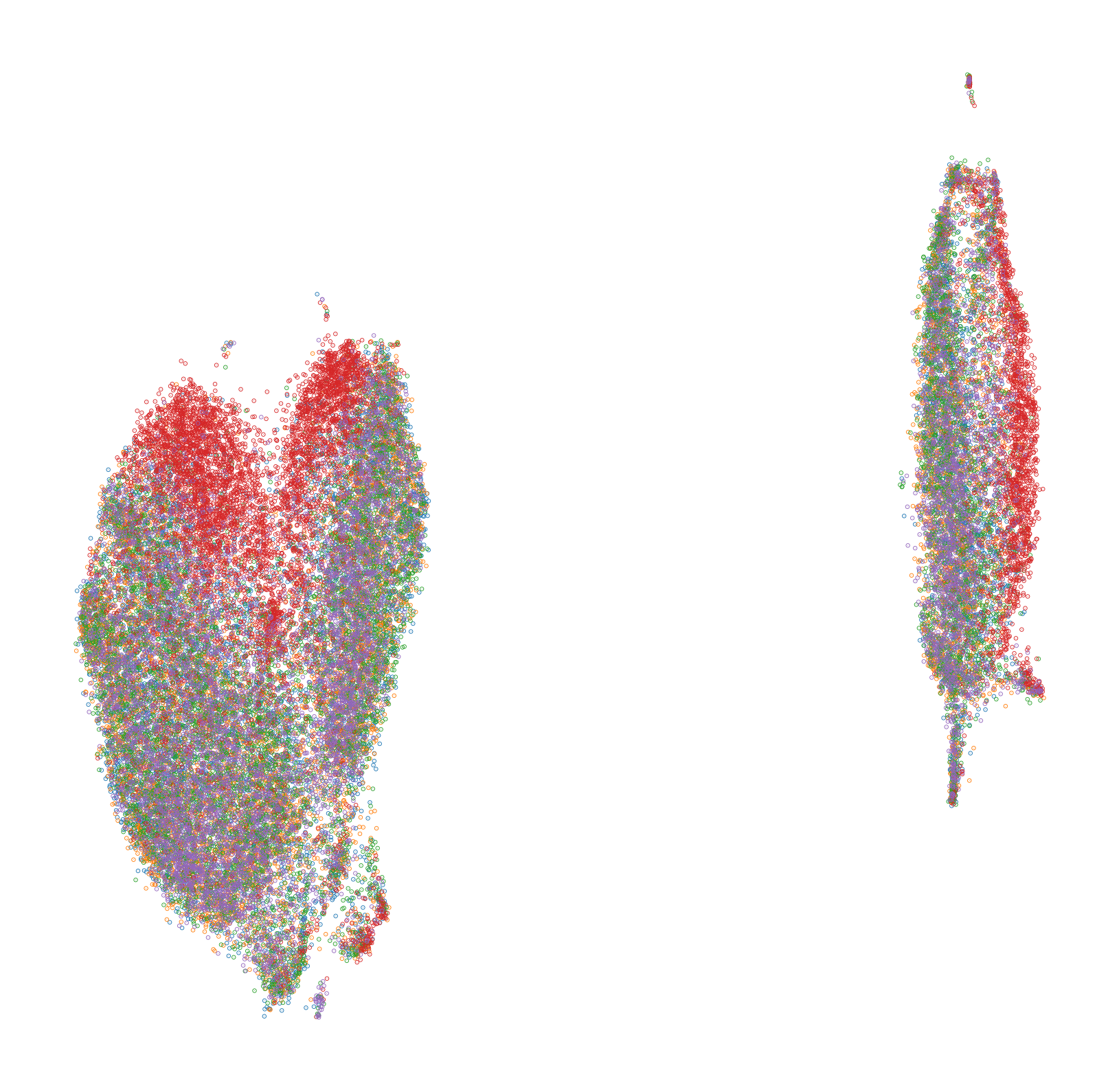}{
    \label{t5-emb}} % 0.45
    }
    % \vspace{-5pt}
\caption{UMAP projections of pre-trained sentence representations from BART and T5 for ROCStories dataset. Embeddings of the sentences are colored based on their position in the ordered sequence $S^*$. It is easier to identify the gold sentence position from the pre-trained BART embeddings.}
\label{fig:embedding}
\end{figure}

\subsection{BART vs. T5} %\SC{Delete first sentence if you need space}
%Different pre-trained models exploit different self-supervised objective functions. 
We want to study the effect of BART's pre-training objective on its performance in sentence ordering task. % One of the tasks BART is pre-trained for (among others) is the rearrangement of permuted sentences which is closely relevant to our task.
BART is pre-trained on multiple tasks including the rearrangement of permuted sentences, which is closely relevant to our task. To investigate if this pre-training objective provides an edge to BART, we conduct the following experiment on the ROCStories dataset. We visualize the UMAP~\cite{mcinnes2018umap} projections of sentence representations obtained from pre-trained BART-\textit{large} and T5-\textit{large} models, and color code them according to their position in the ordered text $S^*$. For example, red color represents the \emph{first} sentence of every instance. We compare with T5, which has a similar architecture but is not pre-trained with \emph{sentence permutation} objective.

 \begin{table}[t!]
	\begin{center}
		\resizebox{0.47\textwidth}{!}{
        \input{figures/training_ablation}
		}
% 		\vspace{-6pt}
		\caption{ Ablations with training setups on ROCStories.}
		\label{ablation}
	\end{center}
\end{table}

%\SC{Readability comment: Instead of directly referring to the figure, have a sentence that describes it. Figure~\ref{fig:embedding} shows this visualization. We observe that... }  
We observe that in case of BART, sentence embeddings belonging to an identical output position ($s_i$) are better clustered in space, making them easier to be identified as shown in Figure~\ref{fig:embedding}.\footnote{An interesting observation from Figure~\ref{fig:embedding} is that both pre-trained BART and T5 embeddings have two distinct clusters. Upon closer inspection we found out that the smaller cluster on the right in both cases correspond to sentences that appear first in the shuffled input (starting with ``\texttt{<S0>}''). } % Other sentences in the input are part of a larger cluster on the left.
In case of T5, the overlap among embeddings at different sentence positions is higher.  To quantify the overlap we measure cluster purity following \citet{ravfogel2020unsupervised}. We perform $k$-means clustering on UMAP projections of sentence embeddings from pre-trained BART \& T5 models ($k=5$, ROCStories has 5 sentences per instance). We measure average purity in each cluster by computing the relative proportion of the most common sentence position. The mean cluster purity for BART: 35.9\% and T5: 23.6\%. This indicates that since pre-trained BART is already able to segregate sentences based on their original position, it finds it easier to reorder them given a shuffled set.

The impact on downstream performance is shown in Table~\ref{index-vs-text}, where BART outperforms T5 in both setups. We posit that  \emph{sentence permutation} denoising % strategy 
during pre-training gives BART an advantage in the sentence ordering task.
%\fbb{if you want you can shorten it and put is as footnote.}\fbb{this is a good observation, but what does it support or show with respect to our methodology?}

%\SC{This para can go into footnote, if you need space}

%\SC{Things are switching frequently in Sec 4. stuff before 4.1 talks about \rebart then there is an unsupported conclusion about  about text-2-marker (t2m) vs text-2-text (t2t). 4.1 jumps to a different issue of BART vs T5. 4.2 starts with t2m vs t2t. second half of 4.2 talks about BART vs T5 again. I suggest the following: (1) compare\rebart to all published baselines. (2) compare t2m vs t2t (first para of 4.2) (3) provide the two reasons for superior performance of \rebart. (4) Have a new section on BART vs T5. This is talk about 4.1 and second half of 4.2 (it's ok to refer to Table 5 twice). }

\begin{table*}[h!]
	\begin{center}
		\resizebox{\textwidth}{!}{
        \input{figures/transfer-results}
		}
% 		\vspace{-6pt}
		\caption{\small Performance of our model when trained on a dataset and evaluated on another in a zero-shot setup. The best and second-best performance for any metric are in bold and underline respectively.
		$^*$We include the performance of BERSON for comparison purposes, when it is evaluated on the same dataset it is fine-tuned on
		%when fine-tuned on individual datasets 
		(from Table~\ref{results}). 
		%for comparison purposes.
		}
		\label{tab:transfer}
	\end{center}
\end{table*}

\subsection{Ablations}
\label{sec:ablations}
We perform a series of ablation experiments with different setups to better understand the working of our model. All the experiments in this section were performed on ROCStories dataset. %\SC{Start each ablation experiment with a new para} 

In the first ablation test, we want to verify whether the model is able to capture coherence among sentences or is just over-fitting on the data. To this end, we train our model using an arbitrarily shuffled order as output instead of the ground-truth order. We observe near random prediction performance as shown in the second row of Table~\ref{ablation} %\SC{If possible make the table appear here. Also, text refers to Table 6 before 5. It would make sense to swap them. }. 
%instead of the ground-truth order, we provide an arbitrarily shuffled order as output
%we provide an arbitrarily shuffle the output of each input instance and observe near random prediction performance (second row in Table~\ref{ablation}).

Next, we examine whether the sentence-markers provide any strong supervision to the model during training. Our initial assumption was that the model can use these markers adequately to learn sentence ordering. To validate our assumption, we remove the sentence-markers from the input (the input is simply a sequence of shuffled sentences) and evaluate if the model is implicitly able to figure out the sentence positions. We observe a significant drop in $\tau$ (-14.97\%) and PMR (-6.19\%) comparing the third and last row in Table~\ref{ablation}
%\SC{One question that you might want to be prepared for: the input has no markers and the output still requires the model to output markers. Yet the performance is quite good ~84\%. What is the reason behind this? Do you think you would get similar (small) drop in performance on other datasets with longer sentences and longer inputs?}
. This result shows that sentence-markers are indeed helpful.

 Finally, we investigate if the sequential nature of sentence markers have an impact on the performance. We append every sentence in an input with \textit{random sentence markers} between 0-100 (e.g. \texttt{\textless S47\textgreater, \textless S78\textgreater}\; etc.). We observe that the model performance is quite close to the final setup (fourth row in Table~\ref{ablation}). There is a slight drop in performance which can be attributed to inconsistent assignment of sentence markers across different instances. % From this result, it is clear that 
This shows that the model can still effectively exploit sentence markers and their sequential nature have little impact on the final performance.

\subsection{Zero-shot Performance}
% \subsection{Transferring Across Datasets}

We investigate how well our model is able to generalize across different datasets. To this end, we evaluate the zero-shot performance of our model on different datasets. 
%Transfer without fine-tuning on downstream NLP datasets has shown to perform poorly~\cite{bowman-etal-2015-large}. \sbrc{should I remove this? any suggestions to rephrase?}

% Table~\ref{tab:transfer} shows the experiment results where 
In our experiment, we train the \rebart model on a single dataset and test it on all others in a zero-shot setup.\footnote{We do not report the results of zero-shot experiments for the arXiv dataset because the training data in arXiv may overlap with NeurIPS, AAN and NSF abstract test sets.} From the results in Table~\ref{tab:transfer}, we observe that in most zero-shot setups \rebart % \SC{nomenclature inconsistency: what does model refer to? Is it "framework + underlying s2s model" or is it "framework + underlying s2s model" trained on a particular dataset? I think you should use "model" to refer to the first thing only and rephrase the sentence to avoid ambiguous meaning of "model"} 
is able to perform well across different domains. %Particularly, \rebart fine-tuned on Wikipedia Movie Plots generalizes well to other unseen datasets, and surprisingly it outperforms the previous state-of-the-art BERSON on PMR score for all datasets except NSF abstract (see the last row for comparison). 
Particularly, \rebart fine-tuned on Wikipedia Movie Plots generalizes well to other unseen datasets. Surprisingly it even outperforms the previous state-of-the-art BERSON, which was fine-tuned on in-domain data, on PMR score for all datasets except NSF abstract (see the last row for comparison). We posit that the presence of longer sentences with more complex language in the Movie Plots dataset  helps the model generalize to other datasets.

\rebart trained on ROCStories performs the worst across all datasets. Its poor performance can be attributed to the fact that ROCStories has fixed length stories with short sentences and simpler language, which makes transfer to other complex datasets harder. %\SC{Since you do not prove it anywhere, it would be helpful to just say "features that could have been easily". BTW, before this, talk about how the sentences are short and the language is simpler which might make transfer to datasets with more complex sentences like ... difficult. To me, this sounds like a more plausible reason. We know that RoCStories are biased but only the last sentence. Also, we don't know if the other datasets are not biased.  }  easily picked up by the model, preventing it from generalization on out-of-domain datasets. 
However, it performs reasonably well on SIND, where the data is from a similar domain %\SC{When you say "a similar \textbf{story} domain", you hint that there are different sub-categories of story-domains and RocStories and SIND are from the same sub-category. You might want to drop the word "story" altogether} 
and most instances are five-sentence long.

 \begin{figure}[t!]
    \centering
    \input{figures/complexity-wise-performance}
    % \vspace{-6pt}
    \caption{ Variation in performance metrics %(a) Accuracy (b) PMR and (c) Kendall's tau ($\tau$) 
    with relative degree of shuffling $\hat{d}(S', S^*)$ % across all datasets
    . A decline in performance is observed with a higher degree of shuffling across datasets.}
    \label{fig:complexity-wise}
\end{figure}
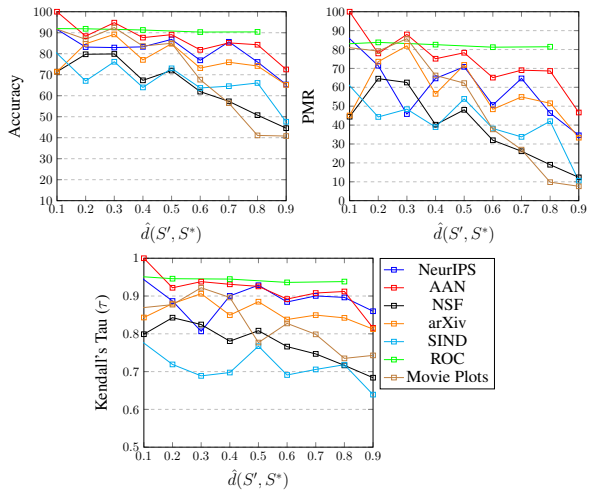

From the results in Table~\ref{tab:transfer}, we also observe that \rebart %\SC{I am confused: are we talking about fine-tuned (on the same dataset for test and train) or about zero-shot setup?} \fbb{it's still zero-shot. Is it more clear now?}\sout{perform}
performs equally well across domains (narrative $\rightarrow$ abstract, abstract $\rightarrow$ narrative). The model trained on Wikipedia Movie Plots (narrative domain), achieve the best zero-shot performance on AAN and NSF abstract (abstract domain).
We also observe good performance during (narrative $\rightarrow$ abstract) transfer, when  %\SC{This is confusing: my understanding is that best zero-shot performance on rocstories is when trained on wiki data and not on AAN and NSF. Maybe you need to elaborate here. } 
 {\rebart} trained on AAN and NSF is tested on ROCStories.  From these experiments, we show that our model is able to generalize across domains and is not restricted to the domain of the dataset it is trained on.

\section{Analysis}
In this section, we perform experiments to explore \rebart's functioning with variation in different aspects of inputs.
%\SC{You can shorten this to keep only one of the two parts of this sentence.}

\subsection{Effect of Shuffling} 
We analyze if \rebart's performance is sensitive to the degree of shuffling in the input. To this end, we define the degree of shuffling $d(S', S^*)$ as the minimum number of swaps required to reconstruct the ordered sequence $S^*$ from $S'$. Lower $d(S', S^*)$ indicates that the input $S'$ is more similar to the ordered output sequence $S^*$. To effectively compare the performance across all datasets, we compute the normalized %\SC{Why do you name it "relative" instead of "normalized"?}
degree of shuffling as: $$\hat{d}(S', S^*) = \dfrac{d(S', S^*)}{|S^*|}$$

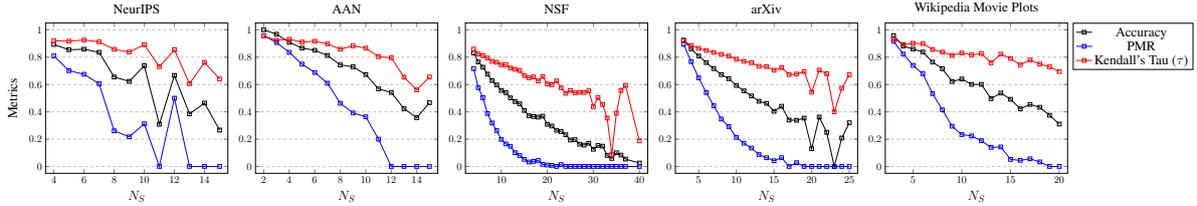
\begin{figure*}[t!]
    \centering
    \input{figures/num_sents_wise-perf}
    % \vspace{-8pt}
    \caption{ Variation of performance metrics (Accuracy, PMR and Kendall's Tau) with the number of input sentences $N_S$ across different datasets. A decline in performance is observed when the number of input sentences increases.}
    \label{var-with-ns}
\end{figure*}

In Figure~\ref{fig:complexity-wise}, %\SC{Fig 5 is referred to before 4. Switch their positions}, 
we observe a gradual decline in performance across all metrics with an increase in the normalized degree of shuffling, $\hat{d}(S', S^*)$. %\sout{The reduction in performance is less profound for NeurIPS, AAN and ROCStories datasets.} %\SC{Provide a reason for this observation} \sbrc{don't have one, avg number of sentences could've been one but SIND also has similar sentence count yet shows decline} 
%\SC{Can the reason of this observation be simpler language? Don't know, just asking. Have you looked at vocabulary sizes?} 
%\SC{Since you don't have a reason, remove this sentence}
Overall, the results show that \rebart performance is higher when $\hat{d}(S', S^*)$ is lower. This could be because a lower degree of shuffling means more coherent and meaningful input, resulting in an easier task for the model.
%\sout{context for individual tokens, thereby solving an easier task.} % \SC{so far you have been using  input instead of context?}

\subsection{Effect of Input Length}
In this experiment, we analyse how \rebart performance varies with the number of sentences in the input. Figure~\ref{var-with-ns} shows \rebart's performance for inputs with different number of sentences, $N_S$. We observe a general declining trend in performance with increasing input length across different datasets.\footnote{We do not show the results on ROC and SIND, because these datasets mostly %\SC{why "mostly"?}\sbrc{it varies a bit for SIND} 
have a fixed number of input sentences.} This shows that the model finds it difficult to tackle longer input instances. % \sbrc{I  didn't  find the sentence}\SC{This "absolute" sentence is misleading. Get rid of this sentence} 
The drop in performance is more pronounced for NSF and arXiv which have instances with higher number of sentences compared to other datasets. For all datasets, we observe that the rate of decline in $\tau$ is much less than Accuracy and PMR. From this observation, we infer that even if the predicted positions of individual sentences are incorrect, our model produces sentence order which are correlated with the original order.

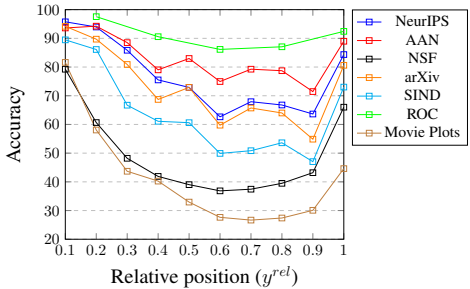
\begin{figure}[t!]
    \centering
    \resizebox{0.4\textwidth}{!}{
    \input{figures/position_wise}
    }
    % \vspace{-5pt}
    \caption{ Position-wise accuracy. A higher prediction accuracy is observed for first and last sentences across all datasets.}
    \label{fig:pos-wise}
\end{figure}

\subsection{Position-wise Performance}

Here, we explore how the performance of \rebart varies while predicting sentences at different positions in the ordered output. To uniformly investigate this across all datasets, we measure performance using a \emph{relative output position} defined as $y^{rel} = \frac{y_i}{\lvert S \rvert}$.
%\noindent where $y^{rel} \in (0, 1]$. 
We consider $y^{rel}$ correct to 1 decimal place and compute the prediction accuracy for each $y^{rel}$. The position-wise prediction accuracy for all datasets is shown in Figure~\ref{fig:pos-wise}. We observe that prediction accuracy is the highest for the first sentence, then there is a steady decline till it starts to rise again towards the end of the output sequence.

 We conjecture that \rebart is able to pick up on shallow stylistic cues, often present in the first and last sentences enabling it to have higher prediction accuracies for these positions. %For instance, 
 For example, in ROCStories all first sentences have a proper noun and introduce the protagonist of the story. In the abstracts, many papers start with similar phrases like ``In this paper,'', ``We present '' %etc. Ending sentences often have distinct patterns like 
 and ends with ``Our contributions are '', ``We achieve '', etc. For Movie plots, last sentence accuracy is significantly less than other datasets because we consider the first 20 sentences only.

\begin{table}[t!]
        \centering
		\resizebox{0.43\textwidth}{!}{
        \input{figures/acc_first_last}
		}
% 		\vspace{-6pt}
		\caption{Accuracy of predicting the first and the last sentences on arXiv and SIND datasets. \rebart achieves the best performance for both datasets.}
		\label{tab:head-tail-acc}
\end{table}

Following previous works \cite{gong2016end, cui2018deep}, we report the prediction accuracy of the head and tail (first and last) sentences for arXiv and SIND in Table~\ref{tab:head-tail-acc}. \rebart outperforms all baselines by a large margin on both datasets. %Further analysis on {\rebart}'s predictions and how performance is impacted with other input factors is presented in the Appendix. %~\ref{sec:appendix}. 

\begin{figure*}[t!]
    \centering
    \includegraphics[width=\textwidth, keepaspectratio]{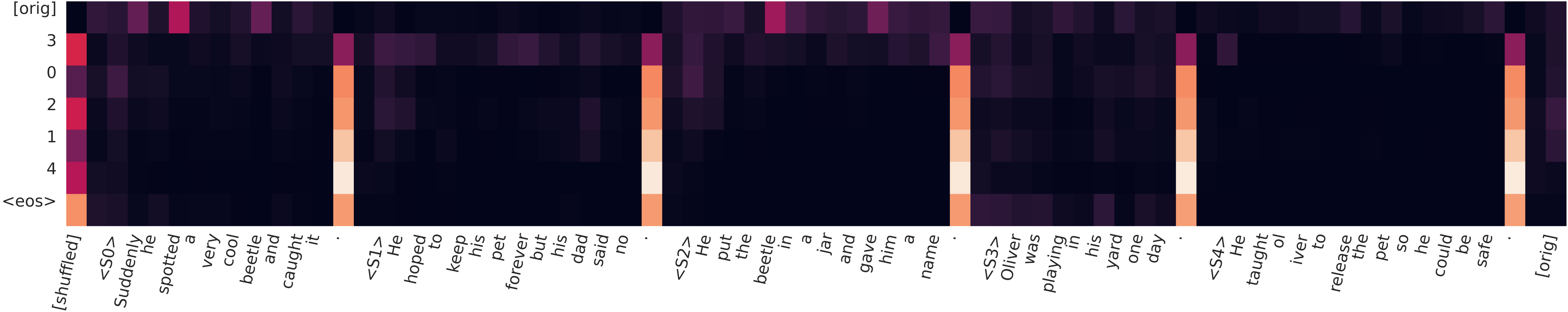}
    % \vspace{-20pt}
    \caption{Visualization of cross-attention in the trained {\rebart} model for an input instance from ROCStories. The $y$-axis shows output tokens, $x$-axis shows input tokens, and colorized cells denote the cross-attention between tokens at a position $(x, y)$. Lighter color indicates higher attention values. The model learns to attend around sentence markers and other special tokens.}
    \label{fig:attn}
\end{figure*}

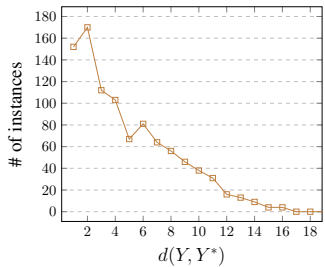
\begin{figure}[t!]
    \centering
    \input{figures/degree_correctly_predicted}
    \caption{ Plot shows how many instances were incorrectly predicted for each $d(Y, Y^*)$ for %Wikipedia 
    Movie Plots.}
    \label{fig:var-acc-dyy}
\end{figure}

\subsection{Prediction Displacement}

For instances where the model prediction was wrong $Y \neq Y^*$, we investigate how far was the model prediction $Y$ was from the gold label $Y^{*}$. To evaluate this we compute $d(Y, Y^*)$, the minimum number of swaps required to retrieve $Y^*$ from $Y$. We experiment on Wikipedia Movie Plots dataset where the performance of \rebart was not as good as other datasets. From Figure~\ref{fig:var-acc-dyy}, we observe that most of the incorrectly predicted samples had a low $d(Y, Y^*)$, with 70\% of the incorrect predictions having $d(Y, Y^*) \leq 6$. This shows that even if the model make a wrong prediction, it mostly misses a few positions and does not get the entire order wrong. %\SC{Why is accuracy not measuring this phenomenon?} \sbrc{it does but I feel quantization of accuracy is not every elegant, which $d_{s}(Y, Y^*)$ handles better} \SC{You need to justify defining a new measure in the text. }

\subsection{Attention Visualization}
\label{sec:attention}
We visualize the norm-based cross-attention map \cite{kobayashi2020attention}, between the decoded output and encoder input, of one of the attention head in Figure~\ref{fig:attn}. Lighter color indicates higher attention values. We append all input instances with special tokens \texttt{[shuffled]} and \texttt{[orig]} %\SC{it seems like you do this for this particular experiment only.} 
at the beginning and end respectively, along with sentence markers at the start of each sentence. In Figure~\ref{fig:attn}, we observe that the model attends to tokens near those special tokens. This shows that during decoding the model finds only tokens next to the sentence markers useful. We hypothesize this is due to the fact that these tokens are able to encapsulate the context of the corresponding sentence. We observe similar maps across different attention heads.

\subsection{Effect of Sentence Displacement} 
%\SC{why did you move this experiment to appendix? You have space, right?}
We investigate if there is any variation in performance if a sentence is placed too far from its position in the shuffled sentence. We compute relative distance from the original position $\delta^{rel}(s_i)$ as:
$$\delta^{rel}(s_i) = \frac{|i - j|}{|S^*|} \; \mathrm{s.t.}\;  s_i = s'_j$$

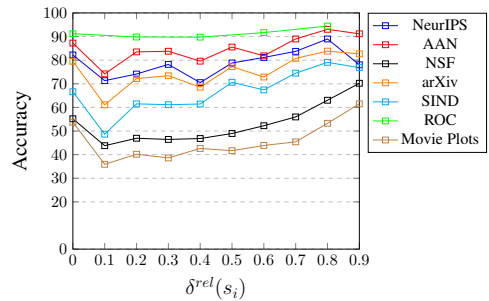
\begin{figure}[t!]
    \centering
\input{figures/acc_wrt_distance_from_orig_position}
    \caption{ Accuracy at a position based on how far it is from the original position. Accuracy doesn't change much with $\delta^{rel}(s_i)$.}
    \vspace{-8pt}
    \label{fig:dist-wise}
\end{figure}

Figure~\ref{fig:dist-wise} shows how the performance varies with respect to $\delta^{rel}(s_i)$. We observe that accuracy doesn't change much with relative displacement. %\SC{I think there is a pattern: acc is increasing slightly towards the end but maybe you can not justify that. In any case, instead of saying that you don't observe a pattern, say that accuracy does not change much (unchanging accuracy is a pattern)}  
We infer that local sentence-level relative displacement doesn't dictate the performance as much as global input-level factors like degree of shuffling and input length.

\section{Conclusion}

In this work, we address the task of sentence ordering by formulating it as a conditional text generation problem. %\SC{can get rid of this sentence if needed} 
We observe that simply generating output text from shuffled input sequences is difficult due to neural text degeneration. 
We solve this problem by proposing  {\rebart}, a text-to-marker generation framework. {\rebart} achieves the state-of-the-art performance on {7} benchmark datasets and is able to generalize well across different domains in a zero-shot setup. %\SC{If you run out of space, you can get rid of this sentence about BART and T5} We found that BART is a better choice over T5 as the underlying text generation model in our framework possibly due to its pre-training objective. 
We investigated the limitations of our model, and found that {\rebart} is sensitive to various factors like number of input sentences and degree of shuffling. Future works can focus on developing models which are robust to such factors.

\clearpage
% \SC{Start references from a fresh page}
\bibliography{emnlp2020}
\bibliographystyle{acl_natbib}

% comment out for submission
% \newpage
% \appendix
% \input{appendix}
% \label{sec:appendix}

\end{document}

%% file: figures/dataset_stat.tex
		\begin{tabular}{ l c c c p{0.1cm} c c p{0.1cm} c c} % <-- Alignments: 1st column left, 2nd middle and 3rd right, with vertical lines in between
			\toprule
			\multirow{2}{*}{Dataset} & \multicolumn{3}{c}{Split} & & \multicolumn{2}{c}{Length} & & \multicolumn{2}{c}{Tokens / sentence} \\ %\TBstrut\\
% 			\cline{2-8}
            \hhline{~---~--~--}
			& {Train} & {Dev} & {Test} & & {Max} & {Avg} & & {Max} & {Avg} \TBstrut\\
			\hline
			NeurIPS  & 2.4K & 0.4K & 0.4K & & 15 & 6 & & 158 & 24.4\Tstrut\\
			AAN  & 8.5K & 962 & 2.6K & & 20 & 5 & & 543 & 20.7\\
			NSF  & 96K & 10K & 21.5K & & 40 & 8.9 & & 307 & 24.3 \\
			arXiv  & 885K & 110K & 110K & & 35 & 5.4 & & 443 & 23.6\\ \hline
			ROCStories & 78K & 9816 & 9816 & & 5 & 5 & & 21 & 9.1\Tstrut\\
			SIND & 78K & 9.8K & 9.8K & & 5 & 5 & & 137 & 10.7\\
			Movie Plots & 27.9K & 3.5K & 3.5K & & 20 & 13.5 & & 319 & 20.4\\
			\bottomrule
		\end{tabular}

% NeurIPS  & 2,427 & 408 & 377 & & 15 & 6 & & 158 & 24.4\Tstrut\\
% AAN  & 8,569 & 962 & 2626 & & 20 & 5 & & 543 & 20.7\\
% NSF  & 96,070 & 10,185 & 21,580 & & 40 & 8.9 & & 307 & 24.3 \\
% arXiv  & 885,216 & 110,726 & 110,669 & & 35 & 5.4 & & 443 & 23.6\\ \hline
% ROCStories & 78,530 & 9816 & 9816 & & 5 & 5 & & 21 & 9.1\Tstrut\\
% SIND & 78,530 & 9816 & 9816 & & 5 & 5 & & 137 & 10.7\\
% Movie Plots & 27,910 & 3488 & 3488 & & 20 & 13.5 & & 319 & 20.4\\

%% file: figures/hyperparams.tex
\begin{tabular}{ l c c c} % <-- Alignments: 1st column left, 2nd middle and 3rd right, with vertical lines in between
	\toprule[1pt]
	{Dataset}  & epochs & learning rate & batch size \TBstrut\\
	\midrule[1pt]
	NeurIPS abstract & 10 & 5e-6  & 4\\
	AAN abstract & 5 & 5e-6 & 4  \\
	NSF abstract & 3 & 5e-6 & 2\\
	arXiv abstract & 3 & 5e-6 & 2\\
	ROCStories & 5 & 5e-6 & 4\\
	SIND & 5 & 5e-6 & 4 \\
	Wikipedia Movie Plots & 5 & 5e-6 & 2\\
	\bottomrule[1pt]
\end{tabular}

%% file: figures/results.tex
\renewcommand{\arraystretch}{1.2}
\begin{tabular}{ l |c c c|c c c|c c c| c c c} % <-- Alignments: 1st column left, 2nd middle and 3rd right, with vertical lines in between
	\toprule[1pt]
	\multirow{2}{*}{\textsc{Method}} & \multicolumn{3}{c}{\textbf{NeurIPS abstract}} & \multicolumn{3}{c}{\textbf{AAN abstract}} & \multicolumn{3}{c}{\textbf{NSF abstract}} & \multicolumn{3}{c}{\textbf{arXiv abstract}}  \TBstrut\\
	%\cline{2-3}
	 & Acc & {PMR} & {$\mathbf{\tau}$} & Acc & {PMR} & {$\mathbf{\tau}$} & Acc & {PMR} & {$\mathbf{\tau}$} & Acc & {PMR} & {$\mathbf{\tau}$}  \TBstrut\\
	\midrule[1pt]
	Pairwise Model \cite{chen2016neural} & - & - & - & - & - & - & - & - & - & - & 33.43 & 0.66  \Tstrut\\
	LSTM+PtrNet \cite{gong2016end} & 50.87 & - & 0.67 & 58.20 & - & 0.69 & 32.45 & - & 0.52 & - & 40.44 & 0.72  \\
	V-LSTM+PtrNet \cite{logeswaran2018sentence} & 51.55 & - & 0.72 & 58.06 & - & 0.73 & 28.33 & - & 0.51 & - & - & -   \\
	ATTOrderNet \cite{cui2018deep} & 56.09 & - & 0.72 & 63.24 & - & 0.73 &  37.72 & - & 0.55 &  - & 42.19 & 0.73 \\
	HAN \cite{wang2019hierarchical} & - & - & - & - & - & - & - & - & - & - & 44.55 & 0.75  \\
	SE-Graph \cite{yin2019graph} & 57.27 & - & 0.75 & 64.64 & - & 0.78 & - & - & - & - & 44.33 & 0.75  \\
	FUDecoder \cite{yin2020enhancing}  & - & - & - & - & - & - & - & - & -&  - & 46.58 & 0.77  \\
	TGCM \cite{oh2019topic} & 59.43 & 31.44 & 0.75 & 65.16 & 36.69 & 0.75 & 42.67 & 22.35 & 0.55 & 58.31 & 44.28 & 0.75  \\
	RankTxNet \cite{kumar2020deep} & - & 24.13 & 0.75 & - & 39.18 & 0.77 & - & 9.78 & 0.58 & - & 43.44 & 0.77 \\
	B-TSort \cite{prabhumoye2020topological} & 61.48 & 32.59 & 0.81 & 69.22 & 50.76 & 0.83 &  35.21 & 10.44 & 0.66 & - & - & -  \\
	BERSON \cite{cui2020bert} & \underline{73.87} & \underline{48.01} & \underline{0.85} & \underline{78.03} & \underline{59.79} & 0.85 & \underline{50.02} & \underline{23.07} & \underline{0.67} & \textbf{75.08} & \underline{56.06} & \underline{0.83}  \Bstrut\\
	\hline
	\textsc{Bart} (fine-tuned) & 64.35 & 33.69 & 0.78 & 73.02 & 52.40 & \underline{0.86} & 33.59 & 14.44 & 0.53 & 60.51 & 2.45 & 0.25  \Tstrut\\
	\textbf{\rebart} &  \textbf{77.41} & \textbf{57.03} & \textbf{0.89} & \textbf{84.28} & \textbf{73.50} & \textbf{0.91} & \textbf{50.23} & \textbf{29.74} &  \textbf{0.76} & \underline{74.28} & \textbf{62.40} & \textbf{0.86}  \\
	%& \small{(+4.7\%)} & \small{(+18.7\%)} & \small{(+4.7\%)} & \small{(+8.0\%)} & \small{(+22.9\%)} & \small{(+7.0\%)} & \small{(-2.6\%)} & \small{(+12.0\%)} & \small{(+8.9\%)} & \small{(+10.3\%)} & \small{(+36.1\%)} & \small{(+10.7\%)} \\
	\bottomrule[1pt]
\end{tabular}

%% file: figures/results-story.tex
		\begin{tabular}{ l |c c c|c c c|c c c} % <-- Alignments: 1st column left, 2nd middle and 3rd right, with vertical lines in between
			\toprule[1pt]
			\multirow{2}{*}{\textsc{Method}} & \multicolumn{3}{c}{\textbf{SIND}} & \multicolumn{3}{c}{\textbf{ROCStories}}  & \multicolumn{3}{c}{\textbf{Wikipedia Movie Plots}} \TBstrut\\
			%\cline{2-3}
			 & Acc & {PMR} & {$\mathbf{\tau}$} & Acc & {PMR} & {$\mathbf{\tau}$} & Acc & {PMR} & {$\mathbf{\tau}$} \TBstrut\\
			\midrule[1pt]
			Pairwise Model \cite{chen2016neural} & - & - & - & - & - & - & - & - & -  \Tstrut\\
			LSTM+PtrNet \cite{gong2016end} & - & 12.34 & 0.48 & - & - & -  & - & - & - \\
			V-LSTM+PtrNet \cite{logeswaran2018sentence} & - & - & - & - & - & -  & - & - & - \\
			ATTOrderNet \cite{cui2018deep} &  - & 14.01 & 0.49 & - & - & - & - & - & - \\
			HAN \cite{wang2019hierarchical} &  - & 15.01 & 0.50 & - &  39.62 & 0.73 & - & - & -\\
			SE-Graph \cite{yin2019graph} & - & 16.22 & 0.52 & - & - & - & - & - & -\\
			FUDecoder \cite{yin2020enhancing} &  - & 17.37 & 0.53 &- & 46.00 & 0.77 & - & - & -\\
			TGCM \cite{oh2019topic} & 38.71 & 15.18 & 0.53 & - & - & -  & - & - & -\\
			RankTxNet \cite{kumar2020deep} & - & 15.48 & 0.57 & - & 38.02 & 0.76 & - & - & -\\
			B-TSort \cite{prabhumoye2020topological}, & 52.23 & 20.32 & 0.60 & - & - & - & - & - & -\\
			BERSON \cite{cui2020bert} &  \underline{58.91} & \underline{31.69} & \underline{0.65} & \underline{82.86} & \underline{68.23} & \underline{0.88} & - & - & - \Bstrut\\
			\hline
			\textsc{Bart} (fine-tuned) & 54.50 & 26.73 & 0.64 & 80.42 & 63.50 & 0.85 & \underline{30.01} & \underline{18.88} & \underline{0.59} \Tstrut\\
			\textbf{\rebart} &  \textbf{64.99} & \textbf{43.15} & \textbf{0.72}  & \textbf{90.78} & \textbf{81.88} & \textbf{0.94} & \textbf{42.04} &  \textbf{25.76} & \textbf{0.77} \\ 
			% &  \small{(+10.3\%)} & \small{(+36.1\%)} & \small{(+10.7\%)} & \small{(+9.5\%)} & \small{(+20.0\%)} & \small{(+6.8\%)} & & & \TBstrut\\
			\bottomrule[1pt]
		\end{tabular}

%% file: figures/text_vs_index.tex
		\begin{tabular}{l|c c c| c c c}
			\toprule[1pt]
			\multirow{2}{*}{\textbf{Setup}} & \multicolumn{3}{c}{\textbf{BART}} & \multicolumn{3}{c}{\textbf{T5}}  \TBstrut\\
			& {Acc} & {PMR} & {$\tau$}   & {Acc} & {PMR} & {$\tau$} \TBstrut\\
			\midrule[1pt]
			text-to-text & 80.42 & 63.50 & 0.85 & 62.75 & 34.49 & 0.71 \Tstrut\\
			text-to-marker  & \textbf{90.78} & \textbf{81.88} & \textbf{0.94} & \textbf{82.36} & \textbf{64.85} & \textbf{0.88} \Tstrut\\
			\bottomrule[1pt]
		\end{tabular}

%% file: figures/training_ablation.tex
		\begin{tabular}{ l |c c} % <-- Alignments: 1st column left, 2nd middle and 3rd right, with vertical lines in between
			\toprule[1pt]
% 			{\textbf{Method}} & \multicolumn{2}{c}{\textbf{ROCStories}} \TBstrut\\
			%\cline{2-3}
			{\textbf{Method}} & {$\mathbf{\tau}$} & {PMR}   \TBstrut\\
			\midrule[1pt]
			Random & 0.00 & 20.00 \Tstrut\\
% 			Ours (\textit{shuffled output}) & 0.0029 & 19.97 \\
            \rebart (\textit{shuffled output}) & 0.00 & 19.97 \\
% 			Ours (\textit{w/o sentence markers}) & 0.7899 & 84.59 \\
            \rebart (\textit{w/o sentence markers}) & 0.79 & 84.59 \\
% 			Ours (\textit{random indices}) & 0.9176 & 88.97 \Bstrut\\
			\rebart (\textit{random markers}) & 0.92 & 88.97 \Bstrut\\
			\hline
% 			Ours (\textit{final setup}) & \textbf{0.9396} & \textbf{90.78} \Tstrut\\
            \rebart (\textit{final setup}) & \textbf{0.94} & \textbf{90.78} \Tstrut\\
			\bottomrule[1pt]
		\end{tabular}

%% file: figures/transfer-results.tex
		\begin{tabular}{ l |c c c| c c c| c c c| c c c| c c c| c c c} % <-- Alignments: 1st column left, 2nd middle and 3rd right, with vertical lines in between
			\toprule[1pt]
			{Evaluated $\rightarrow$} & \multicolumn{3}{c}{\textbf{NeurIPS abstract}} & \multicolumn{3}{c}{\textbf{AAN abstract}} & \multicolumn{3}{c}{\textbf{NSF abstract}} &  \multicolumn{3}{c}{\textbf{SIND}}  & \multicolumn{3}{c}{\textbf{ROCStories}} & \multicolumn{3}{c}{\textbf{Movie Plots}}\TBstrut\\
			%\cline{2-3}
			Trained $\downarrow$ & {Acc} &{PMR} &{$\tau$}   & {Acc} &{PMR} &{$\tau$} & {Acc} &{PMR} &{$\tau$} & {Acc} &{PMR} &{$\tau$} & {Acc} &{PMR} &{$\tau$} & {Acc} &{PMR} &{$\tau$}   \TBstrut\\
			\midrule[1pt]

			\textbf{NeurIPS abstract} &  \textbf{77.41} &57.03 &\textbf{0.89} &  78.42 &55.38 &0.80 & 29.04 &11.37 &0.57 & 55.19 &24.97 &0.59 & 76.99 &58.62 &0.89  & 17.46 &9.39 &0.40\Tstrut\\
			
			\textbf{AAN abstract} &  76.99 &\textbf{58.62} &\underline{0.89} & \textbf{84.28} &\textbf{73.50} &\textbf{0.91} & 37.09 &19.12 &0.62 & 58.08 &33.93 &0.64  & 86.62 &75.46 &0.91 &  24.08 &16.23 &0.48\Tstrut\\
			
			\textbf{NSF abstract} & \underline{77.32} &\underline{57.82} &0.88 & \underline{81.15} &61.10 &0.81 & \textbf{50.23} &\textbf{29.74} & \textbf{0.76} & 57.26 &28.46 &0.60  & 86.62 &75.22 &0.90 &  \underline{30.86} &\underline{17.89} &\underline{0.76}\Tstrut\\
			
			\textbf{SIND} & 59.95 &34.75 &0.77 &  75.36 &53.29 &0.78 & 34.32 &17.27 &0.61 & \textbf{64.99} &\textbf{43.15} &\textbf{0.72} & 86.56 &75.02 &0.91  & 21.49 &13.87 &0.45\Tstrut\\
			
			\textbf{ROCStories} & 15.84 &0.27 &0.12 &  21.43 &0.45 &0.09 & 6.50 & 0.12 &0.07 & 50.09 &21.03 &0.54  &  \textbf{90.78} &\textbf{81.87} &\textbf{0.94} & 2.88 &0.06&0.03\Tstrut\\
			
			\textbf{Movie Plots}  & {73.26} &{54.11} &{0.87} & {78.56} &\underline{64.46} &\underline{0.85} & {39.19} & {20.33} & {0.65} & {58.40} &\underline{36.12} &\underline{0.65} & \underline{87.07} &\underline{76.39} & \underline{0.91} &  \textbf{42.05} &\textbf{25.76} &\textbf{0.77}\TBstrut\\
			\hline
			\textsc{Berson}\textsuperscript{*} & {73.87} &48.01 &0.85 & 78.03 &59.79 &0.85 & \underline{50.02} & \underline{23.07} & \underline{0.67} & \underline{58.91} &31.69 &0.65 & 82.86 &  68.23 &0.88 & - &- &- \TBstrut\\
			\bottomrule[1pt]
		\end{tabular}

%% file: figures/complexity-wise-performance.tex
\begin{tikzpicture}[scale=0.44]
\begin{axis}[
    xlabel={\Large $\hat{d}(S', S^*)$},
    ylabel={\Large Accuracy},
    xmin=0.1, xmax=0.9,
    ymin=10, ymax=100,
    xtick={0.1, 0.2, 0.3, 0.4, 0.5, 0.6, 0.7, 0.8, 0.9},
    ytick={10, 20, 30, 40, 50, 60, 70, 80, 90, 100},
    ymajorgrids=true,
    grid style=dashed,
    legend pos=outer north east
]

% NIPS
\addplot[
    color=blue,
    mark=square,
    ]
    coordinates {
    (0.0, 100.0)
(0.2, 83.17460317460318)
(0.3, 82.93650793650792)
(0.4, 83.27731092436976)
(0.5, 86.82291666666666)
(0.6, 76.87505883708414)
(0.7, 85.68245480010187)
(0.8, 76.06280193236714)
(0.9, 65.33494283494285)
    };
    
    % AAN
    \addplot[
    color=red,
    mark=square,
    ]
    coordinates {
    (0.0, 97.1244131455399)
(0.1, 100.0)
(0.2, 88.37837837837837)
(0.3, 94.7905477980666)
(0.4, 87.67009464192562)
(0.5, 89.0863577636834)
(0.6, 81.83760886809665)
(0.7, 85.16253663735677)
(0.8, 84.28229094138186)
(0.9, 72.50312650312651)
    };
    
    % NSF
     \addplot[
    color=black,
    mark=square,
    ]
    coordinates {
    (0.0, 97.72025984571663)
(0.1, 71.42857142857143)
(0.2, 79.74163319946453)
(0.3, 79.89199803632788)
(0.4, 67.32627873749066)
(0.5, 71.99331619588422)
(0.6, 61.94228987662972)
(0.7, 57.31890438784889)
(0.8, 50.73125805338269)
(0.9, 44.537809434108927)
% (1.0, 0.17684833094950617)
    };

    % Arxiv
    \addplot[
    color=orange,
    mark=square,
    ]
    coordinates {
        (0.0, 94.9192676287427)
        (0.1, 71.4285714285714)
        (0.2, 84.73412335985437)
        (0.3, 89.22255746223297)
        (0.4, 77.05629284746108)
        (0.5, 84.76116312675896)
        (0.6, 73.1627911732666)
        (0.7, 75.935628196322)
        (0.8, 74.15667808939449)
        (0.9, 65.17275620755505)
    };
    
    % SIND
    \addplot[
    color=cyan,
    mark=square,
    ]
    coordinates {
    (0.0, 93.21537789427697)
    (0.2, 67.04207920792075)
    (0.3, 76.17187499999999)
    (0.4, 63.94977324263059)
    (0.5, 73.03864168618269)
    (0.6, 63.69607145278321)
    (0.7, 64.54784408160612)
    (0.8, 66.07019464121736)
    (0.9, 47.57727652464494)
    };

    % ROC
    \addplot[
    color=green,
    mark=square,
    ]
    coordinates {
    (0.0, 92.05882352941178)
    (0.2, 91.87739463601541)
    (0.4, 91.30744567153504)
    (0.6, 90.31795362180203)
    (0.8, 90.39999999999986)
    };
    
    %Wiki Movie Plots
    \addplot[
    color=brown,
    mark=square,
    ]
    coordinates {
    (0.0, 96.55172413793103)
    (0.2, 86.875)
    (0.3, 92.46031746031746)
    (0.4, 83.6584432969975)
    (0.5, 84.99057659841975)
    (0.6, 67.64792791863924)
    (0.7, 56.46590982383009)
    (0.8, 41.095369329986184)
    (0.9, 40.78737355270298)
    };
    %\legend{NeurIPS, AAN, NSF, SIND, ROC, Movie Plots}
\end{axis}
\label{position-wise}
\end{tikzpicture}%% Acc
\begin{tikzpicture}[scale=0.44]
\begin{axis}[
    xlabel={\Large $\hat{d}(S', S^*)$},
    ylabel={\Large PMR},
    xmin=0.1, xmax=0.9,
    ymin=0, ymax=100,
    xtick={0.1, 0.2, 0.3, 0.4, 0.5, 0.6, 0.7, 0.8, 0.9},
    ytick={0, 10, 20, 30, 40, 50, 60, 70, 80, 90, 100},
    ymajorgrids=true,
    grid style=dashed,
    legend pos=outer north east
]

% NIPS
\addplot[
    color=blue,
    mark=square,
    ]
    coordinates {
    (0.0, 100.0)
    (0.2, 71.42857142857143)
    (0.3, 45.83333333333333)
    (0.4, 64.70588235294117)
    (0.5, 70.83333333333334)
    (0.6, 50.63291139240506)
    (0.7, 64.70588235294117)
    (0.8, 46.3768115942029)
    (0.9, 34.61538461538461)
    };
    
    % AAN
    \addplot[
    color=red,
    mark=square,
    ]
    coordinates {
    (0.0, 83.09859154929578)
    (0.1, 100.0)
    (0.2, 77.92792792792793)
    (0.3, 87.96992481203007)
    (0.4, 75.11737089201877)
    (0.5, 78.29457364341084)
    (0.6, 65.1219512195122)
    (0.7, 69.06474820143885)
    (0.8, 68.63636363636364)
    (0.9, 46.666666666666664)
    };
    
    % NSF
     \addplot[
    color=black,
    mark=square,
    ]
    coordinates {
(0.0, 68.81851400730816)
(0.1, 44.44444444444444)
(0.2, 64.57831325301204)
(0.3, 62.54295532646048)
(0.4, 40.20501138952164)
(0.5, 48.08935920047031)
(0.6, 31.98003327787022)
(0.7, 26.16131268777444)
(0.8, 19.01850627891606)
(0.9, 12.342271293375394)
    % (1.0, 0.0)
    };
    
    % arXiv
     \addplot[
    color=orange,
    mark=square,
    ]
    coordinates {
    (0.0, 90.26736771450945)
    (0.1, 45.09803921568628)
    (0.2, 73.56425251162094)
    (0.3, 81.9347539128795)
    (0.4, 56.54274594014226)
    (0.5, 71.85640618802063)
    (0.6, 48.36977368622939)
    (0.7, 54.89692073650527)
    (0.8, 51.55676637384633)
    (0.9, 33.33333333333333)
    };
    
    % SIND
    \addplot[
    color=cyan,
    mark=square,
    ]
    coordinates {
    (0.0, 77.37003058103976)
    (0.2, 44.306930693069305)
    (0.3, 48.4375)
    (0.4, 38.952380952380956)
    (0.5, 53.86416861826698)
    (0.6, 38.18791946308725)
    (0.7, 33.762057877813504)
    (0.8, 42.04545454545455)
    (0.9, 10.526315789473683)
    };

    % ROC
    \addplot[
    color=green,
    mark=square,
    ]
    coordinates {
    (0.0, 82.35294117647058)
    (0.2, 83.78033205619413)
    (0.4, 82.57926612041325)
    (0.6, 81.23356442744442)
    (0.8, 81.46835443037975)

    };
    
    %Wiki Movie Plots
    \addplot[
    color=brown,
    mark=square,
    ]
    coordinates {
    (0.0, 82.75862068965517)
    (0.2, 79.16666666666666)
    (0.3, 85.71428571428571)
    (0.4, 66.26506024096386)
    (0.5, 62.22222222222222)
    (0.6, 37.75100401606426)
    (0.7, 27.010309278350515)
    (0.8, 9.797297297297296)
    (0.9, 7.615230460921844)
    };
    %\legend{NeurIPS, AAN, NSF, SIND, ROC, Movie Plots}
\end{axis}
\label{position-wise}
\end{tikzpicture} %% PMR
\begin{tikzpicture}[scale=0.44] %% Tau
\begin{axis}[
    xlabel={\Large $\hat{d}(S', S^*)$},
    ylabel={\Large Kendall's Tau ($\tau$)},
    xmin=0.1, xmax=0.9,
    ymin=0.5, ymax=1,
    xtick={0.1, 0.2, 0.3, 0.4, 0.5, 0.6, 0.7, 0.8, 0.9},
    ytick={0.5, 0.6, 0.7, 0.8, 0.9, 1.0},
    ymajorgrids=true,
    grid style=dashed,
    legend pos=outer north east
]

% NIPS
\addplot[
    color=blue,
    mark=square,
    ]
    coordinates {
    (0.0, 1.0)
(0.2, 0.8867724867724869)
(0.3, 0.8065519319595408)
(0.4, 0.9002801120448178)
(0.5, 0.928637566137566)
(0.6, 0.8842548450143386)
(0.7, 0.9003006403422005)
(0.8, 0.8962670180061484)
(0.9, 0.8598010232146177)
    };
    
    % AAN
    \addplot[
    color=red,
    mark=square,
    ]
    coordinates {
    (0.0, 0.8961739877790258)
(0.1, 1.0)
(0.2, 0.921793221793222)
(0.3, 0.9380000207313054)
(0.4, 0.9310306282137266)
(0.5, 0.9255050505050513)
(0.6, 0.8919333235939874)
(0.7, 0.9079476935748438)
(0.8, 0.9120174774720227)
(0.9, 0.8156245369562712)
    };
    
    % NSF
     \addplot[
    color=black,
    mark=square,
    ]
    coordinates {
    (0.0, 0.9369498382072212)
(0.1, 0.7989417989417988)
(0.2, 0.842691401177911)
(0.3, 0.8245695384192815)
(0.4, 0.7805289037317957)
(0.5, 0.8082233917926324)
(0.6, 0.7661397016813685)
(0.7, 0.7468148655476194)
(0.8, 0.7164577107926987)
(0.9, 0.6838211855754993)
% (1.0, 0.47307120810825815)
    };
    
    % arXiv
     \addplot[
    color=orange,
    mark=square,
    ]
    coordinates {
        (0.0, 0.9190146500691683)
        (0.1, 0.8431372549019612)
        (0.2, 0.8798996541686576)
        (0.3, 0.9061873983072118)
        (0.4, 0.8494571898543541)
        (0.5, 0.8849511964288558)
        (0.6, 0.8378360489950177)
        (0.7, 0.8495029115378543)
        (0.8, 0.8421848304718812)
        (0.9, 0.8124374722214736)
    };
    
    % SIND
    \addplot[
    color=cyan,
    mark=square,
    ]
    coordinates {
    (0.0, 0.8320687628701264)
    (0.2, 0.7189768976897694)
    (0.3, 0.6886805014097812)
    (0.4, 0.6975086923658345)
    (0.5, 0.7673760137617437)
    (0.6, 0.6909913129376195)
    (0.7, 0.7057226012000228)
    (0.8, 0.7182218917446207)
    (0.9, 0.6390901496164654)
    };

    % ROC
    \addplot[
    color=green,
    mark=square,
    ]
    coordinates {
    (0.0, 0.9558823529411761)
    (0.2, 0.9458492975734348)
    (0.4, 0.9448300195181228)
    (0.6, 0.9360267750418407)
    (0.8, 0.9382278481012621)
    };
    
    %Wiki Movie Plots
    \addplot[
    color=brown,
    mark=square,
    ]
    coordinates {
    (0.0, 0.861762655419951)
    (0.2, 0.8771960261528946)
    (0.3, 0.9218682218973367)
    (0.4, 0.8963203463203463)
    (0.5, 0.7768061012474927)
    (0.6, 0.8275443159176316)
    (0.7, 0.7986115294703048)
    (0.8, 0.7350103753772443)
    (0.9, 0.7428874788421125)
    };
    \legend{\Large NeurIPS, \Large AAN, \Large NSF, \Large arXiv, \Large SIND, \Large ROC, \Large Movie Plots}
\end{axis}
\label{position-wise}
\end{tikzpicture}

%% file: figures/num_sents_wise-perf.tex
\begin{tikzpicture}[scale=0.35] %% NeurIPS
\begin{axis}[
    title = \Large NeurIPS,
	ytick={0, .2, .40, .60, .80, 1.00},
	xlabel = \Large $N_S$,
	ylabel= \Large Metrics,
	enlargelimits=0.05,
% 	ybar interval=0.7,
	ymin=0, ymax=1,
	xmin=4, xmax=15,
    grid style=dashed,
    ymajorgrids=true,
    label style={font=\Large},
    xlabel style={font=\Large},
]

%% Accuracy 
\addplot [
    mark = square,
]
	coordinates {
        % (2, 1.0)
        % (3, 1.0)
        (4, 0.8928571428571429)
        (5, 0.8545454545454547)
        (6, 0.8583333333333332)
        (7, 0.8354978354978355)
        (8, 0.654891304347826)
        (9, 0.6231884057971016)
        (10, 0.7374999999999999)
        (11, 0.3090909090909091)
        (12, 0.6666666666666666)
        (13, 0.38461538461538464)
        (14, 0.4642857142857143)
        (15, 0.26666666666666666)
                };

    %% PMR 
    \addplot[
    color = blue,
    mark = square,
    ]
	coordinates {
            % (2, 1.0)
            % (3, 0.6)
            (4, 0.8095238095238095)
            (5, 0.7012987012987013)
            (6, 0.675)
            (7, 0.6060606060606061)
            (8, 0.2608695652173913)
            (9, 0.21739130434782608)
            (10, 0.3125)
            (11, 0.0)
            (12, 0.5)
            (13, 0.0)
            (14, 0.0)
            (15, 0.0)
            };

    %% Tau
    \addplot[
    color = red,
    mark = square,
    ]
    coordinates {
        % (2, 1.0)
        % (3, 0.7217161238900369)
        (4, 0.9206349206349206)
        (5, 0.9168831168831164)
        (6, 0.9258333333333331)
        (7, 0.9119769119769122)
        (8, 0.8571428571428568)
        (9, 0.8378191856452724)
        (10, 0.8909722222222222)
        (11, 0.730909090909091)
        (12, 0.8545454545454545)
        (13, 0.606060606060606)
        (14, 0.7616561768160858)
        (15, 0.6408937593274027)
                };
\end{axis}
\label{acc-num-sents}
\end{tikzpicture}
\begin{tikzpicture}[scale=0.35] %% AAN
\begin{axis}[
    title = {\Large AAN},
	ytick={0, .2, .40, .60, .80, 1.00},
	xlabel = \Large $N_S$,
	enlargelimits=0.05,
% 	ybar interval=0.7,
	ymin=0, ymax=1,
	xmin=2, xmax=15,
    grid style=dashed,
    ymajorgrids=true,
    legend pos=outer north east
]

%% Accuracy 
\addplot [
    mark = square,
]
	coordinates {
        % (1, 1.0)
        (2, 1.0)
        (3, 0.9682107175295184)
        (4, 0.9089430894308943)
        (5, 0.8667838312829527)
        (6, 0.8503184713375803)
        (7, 0.8128258602711159)
        (8, 0.7431972789115646)
        (9, 0.7297297297297298)
        (10, 0.6727272727272727)
        (11, 0.5681818181818181)
        (12, 0.5416666666666666)
        (13, 0.42307692307692313)
        (14, 0.3571428571428571)
        (15, 0.4666666666666666)
                };

    %% PMR 
    \addplot[
    color = blue,
    mark = square,
    ]
	coordinates {
                    % (1, 0.6666666666666666)
        (2, 0.9574468085106383)
        (3, 0.9073569482288828)
        (4, 0.8357723577235773)
        (5, 0.7486818980667839)
        (6, 0.6878980891719745)
        (7, 0.6094890510948905)
        (8, 0.46258503401360546)
        (9, 0.3918918918918919)
        (10, 0.36363636363636365)
        (11, 0.2)
        (12, 0.0)
        (13, 0.0)
        (14, 0.0)
        (15, 0.0)};

    %% Tau
    \addplot[
    color = red,
    mark = square,
    ]
    coordinates {
        % (1, 1.0)
        (2, 0.9549393376553669)
        (3, 0.9234042832017003)
        (4, 0.931165311653117)
        (5, 0.9107205623901599)
        (6, 0.9166312809624919)
        (7, 0.8968876367995617)
        (8, 0.8586540820829001)
        (9, 0.8828828828828826)
        (10, 0.8666666666666668)
        (11, 0.8036363636363637)
        (12, 0.7954545454545453)
        (13, 0.6538461538461537)
        (14, 0.5604395604395604)
        (15, 0.655155882113266)};
\end{axis}
\label{acc-num-sents-aan}
\end{tikzpicture}
\begin{tikzpicture}[scale=0.35] %% NSF
\begin{axis}[
    title = {\Large NSF},
	ytick={0, .2, .4, .6, .8, 1},
	xlabel = \Large $N_S$,
% 	ylabel=PMR,
	enlargelimits=0.05,
% 	ybar interval=0.7,
	ymin=0, ymax=1,
	xmin=4, xmax=40,
    grid style=dashed,
    ymajorgrids=true,
]

% Accuracy
\addplot[
    mark = square,
    ]
	coordinates {
% (1, 1.0)
% (2, 0.9922178988326849)
% (3, 0.9041322314049579)
(4, 0.8304994686503719)
(5, 0.7652868554829357)
(6, 0.7272433828276302)
(7, 0.6760276760276741)
(8, 0.628397917871602)
(9, 0.5970361347949659)
(10, 0.5566181127295768)
(11, 0.5402908003804866)
(12, 0.5019242155121366)
(13, 0.47502448579823653)
(14, 0.45846964740094537)
(15, 0.4103111111111106)
(16, 0.3715503246753247)
(17, 0.36542218363360196)
(18, 0.36129506990434135)
(19, 0.3686909581646426)
(20, 0.30821917808219174)
(21, 0.2960815322438945)
(22, 0.26466275659824023)
(23, 0.25543478260869557)
(24, 0.23219086021505375)
(25, 0.1937499999999999)
(26, 0.19580419580419578)
(27, 0.16280864197530867)
(28, 0.1564039408866995)
(29, 0.16965517241379313)
(30, 0.12564102564102564)
(31, 0.15322580645161288)
(32, 0.1484375)
(33, 0.08080808080808081)
(34, 0.058823529411764705)
(35, 0.09999999999999999)
(36, 0.08333333333333333)
(37, 0.05405405405405406)
(40, 0.025)};
            
\addplot[
    color = blue,
    mark = square,
    ]
	coordinates {
% (1, 0.7021660649819494)
% (2, 0.6964980544747081)
% (3, 0.7801652892561983)
(4, 0.7162592986184909)
(5, 0.5766158315177923)
(6, 0.5035506778566817)
(7, 0.38803418803418804)
(8, 0.31868131868131866)
(9, 0.2624847746650426)
(10, 0.19822672577580747)
(11, 0.16517189835575485)
(12, 0.14742451154529307)
(13, 0.10088148873653281)
(14, 0.08015267175572519)
(15, 0.052)
(16, 0.032467532467532464)
(17, 0.03690036900369004)
(18, 0.04415011037527594)
(19, 0.015384615384615385)
(20, 0.010273972602739725)
(21, 0.007380073800738007)
(22, 0.0)
(23, 0.013888888888888888)
(24, 0.0)
(25, 0.0)
(26, 0.0)
(27, 0.0)
(28, 0.0)
(29, 0.0)
(30, 0.0)
(31, 0.0)
(32, 0.0)
(33, 0.0)
(34, 0.0)
(35, 0.0)
(36, 0.0)
(37, 0.0)
(40, 0.0)};
            
\addplot[
    color = red,
    mark = square,
    ]
	coordinates {
% (1, 1.0)
% (2, 0.8154855422018229)
% (3, 0.8499017076556322)
(4, 0.8592585447645154)
(5, 0.8238350010343486)
(6, 0.81090385123263)
(7, 0.7939366314780033)
(8, 0.7693292173200771)
(9, 0.762710695987211)
(10, 0.7437346754858184)
(11, 0.74407382352629)
(12, 0.7205955287743816)
(13, 0.7112066254845402)
(14, 0.7018980451524613)
(15, 0.6654099291217204)
(16, 0.6487288710747798)
(17, 0.6554241853759739)
(18, 0.6258056181687283)
(19, 0.659471622874054)
(20, 0.6041632403351986)
(21, 0.5983201187136965)
(22, 0.6264937995371084)
(23, 0.5756887970405201)
(24, 0.535769644145017)
(25, 0.5566516351570056)
(26, 0.5390559910609665)
(27, 0.5431241208157815)
(28, 0.5412684838318748)
(29, 0.5539412419050646)
(30, 0.43801846397502114)
(31, 0.5050913541927614)
(32, 0.45356944223740664)
(33, 0.35431920683849844)
(34, 0.08724260675251742)
(35, 0.3904201364792922)
(36, 0.5560481235220895)
(37, 0.5934413158353018)
(40, 0.1878990420479456)};
\end{axis}
\label{pmr-num-sents}
\end{tikzpicture}
\begin{tikzpicture}[scale=0.35] %% arXiv
\begin{axis}[
    title = {\Large arXiv},
	ytick={0, .2, .4, .6, .8, 1},
	xlabel = \Large $N_S$,
% 	ylabel=PMR,
	enlargelimits=0.05,
% 	ybar interval=0.7,
	ymin=0, ymax=1,
	xmin=3, xmax=25,
    grid style=dashed,
    ymajorgrids=true,
]

% Accuracy
\addplot[
    mark = square,
    ]
	coordinates {
    % (2, 0.9858180732484076)
    (3, 0.9253496503496725)
    (4, 0.8610415495477078)
    (5, 0.8088218682932661)
    (6, 0.7608415629025294)
    (7, 0.7163141118516201)
    (8, 0.6727168199737188)
    (9, 0.6423636965334719)
    (10, 0.5930599369085136)
    (11, 0.5534933207234894)
    (12, 0.5163505956722596)
    (13, 0.4772351753258672)
    (14, 0.46209196209196224)
    (15, 0.4040498442367606)
    (16, 0.4410282258064516)
    (17, 0.3396584440227703)
    (18, 0.337962962962963)
    (19, 0.3540669856459329)
    (20, 0.13)
    (21, 0.36190476190476184)
    (22, 0.25)
    (23, 0.0)
    (24, 0.20833333333333334)
    (25, 0.32)
    % (27, 0.4074074074074074)
};
            
\addplot[
    color = blue,
    mark = square,
    ]
	coordinates {
    % (2, 0.9352109872611465)
    (3, 0.893997668997669)
    (4, 0.76772116318291)
    (5, 0.6486501750818932)
    (6, 0.5409331615858022)
    (7, 0.4447986090988119)
    (8, 0.34730617608409986)
    (9, 0.29188384214445273)
    (10, 0.21307714367651276)
    (11, 0.16991764195925443)
    (12, 0.1349380014587892)
    (13, 0.08711217183770883)
    (14, 0.06388206388206388)
    (15, 0.04205607476635514)
    (16, 0.06451612903225806)
    (17, 0.0)
    (18, 0.027777777777777776)
    (19, 0.0)
    (20, 0.0)
    (21, 0.0)
    (22, 0.0)
    (23, 0.0)
    (24, 0.0)
    (25, 0.0)
    % (27, 0.0)
};
            
\addplot[
    color = red,
    mark = square,
    ]
	coordinates {
        % (2, 0.9213473093492301)
        (3, 0.9203009993223769)
        (4, 0.8853110367399468)
        (5, 0.8653391361875833)
        (6, 0.8495752626855976)
        (7, 0.8349038618871657)
        (8, 0.8206636661949283)
        (9, 0.8085044760206931)
        (10, 0.7858865432476401)
        (11, 0.7695068627730716)
        (12, 0.759632903004706)
        (13, 0.7333651881249086)
        (14, 0.731171164441103)
        (15, 0.7044387171217362)
        (16, 0.7232622335670608)
        (17, 0.6725937275583144)
        (18, 0.6765286834020894)
        (19, 0.6951823653296697)
        (20, 0.5435190564778416)
        (21, 0.7053360420917617)
        (22, 0.6792267333195382)
        (23, 0.4)
        (24, 0.572463768115942)
        (25, 0.6715074579925805)
        % (27, 0.8119658119658121)
};
\end{axis}
\label{pmr-num-sents-arxiv}
\end{tikzpicture}
\begin{tikzpicture}[scale=0.35] %% Wiki Movies
\begin{axis}[
    title = {\Large Wikipedia Movie Plots},
	ytick={0, .2, .4, .6, .8, 1},
	xlabel = \Large $N_S$,
	enlargelimits=0.05,
% 	enlargetitle=0.05,
% 	ybar interval=0.7,
	ymin=0, ymax=1,
	xmin=3, xmax=20,
    grid style=dashed,
    ymajorgrids=true,
    legend pos=outer north east,
    legend style={font=\small},
]

% Accuracy
\addplot[
    mark = square,
    ]
	coordinates {
% 	(2, 0.9695121951219512)
(3, 0.9580838323353293)
(4, 0.8813291139240507)
(5, 0.8591715976331362)
(6, 0.838199513381995)
(7, 0.7642857142857143)
(8, 0.7154471544715447)
(9, 0.6220238095238095)
(10, 0.6399999999999999)
(11, 0.6010695187165775)
(12, 0.6009259259259262)
(13, 0.4964221824686943)
(14, 0.5377551020408163)
(15, 0.49195402298850555)
(16, 0.42257462686567165)
(17, 0.4540182270091136)
(18, 0.43333333333333346)
(19, 0.3746654772524532)
(20, 0.31069424964936915)};
            
% PMR
\addplot[
    color = blue,
    mark = square,
    ]
	coordinates {
% 	(2, 0.6951219512195121)
(3, 0.9161676646706587)
(4, 0.8227848101265823)
(5, 0.7396449704142012)
(6, 0.6788321167883211)
(7, 0.5333333333333333)
(8, 0.4146341463414634)
(9, 0.29464285714285715)
(10, 0.23333333333333334)
(11, 0.2235294117647059)
(12, 0.18888888888888888)
(13, 0.13953488372093023)
(14, 0.14285714285714285)
(15, 0.05172413793103448)
(16, 0.04477611940298507)
(17, 0.056338028169014086)
(18, 0.03333333333333333)
(19, 0.0)
(20, 0.0)};
            
\addplot[
    color = red,
    mark = square,
    ]
	coordinates {
% 	(2, 0.7055821711371589)
(3, 0.9334194963523263)
(4, 0.8930722104768287)
(5, 0.9017751479289935)
(6, 0.8992700729927003)
(7, 0.856349206349206)
(8, 0.8371772799093065)
(9, 0.8132527370309655)
(10, 0.83037037037037)
(11, 0.816898395721925)
(12, 0.8269360269360264)
(13, 0.7587683634195258)
(14, 0.8223999853506638)
(15, 0.7902488316281423)
(16, 0.7441364605543711)
(17, 0.7792046396023198)
(18, 0.7496732026143798)
(19, 0.7296539699553679)
(20, 0.6948523987398099)};
\legend{\Large Accuracy, \Large PMR, \Large Kendall's Tau ($\tau$)}
\end{axis}
\label{perf-num-sents-wiki}
\end{tikzpicture}

%% file: figures/position_wise.tex
\begin{tikzpicture}[scale=0.6]
\begin{axis}[
    xlabel={\Large Relative position ($y^{rel}$)},
    ylabel={\Large Accuracy},
    xmin=0.1, xmax=1,
    ymin=20, ymax=100,
    xtick={0.1, 0.2, 0.3, 0.4, 0.5, 0.6, 0.7, 0.8, 0.9, 1.0},
    ytick={20, 30, 40, 50, 60, 70, 80, 90, 100},
    ymajorgrids=true,
    grid style=dashed,
    legend pos=outer north east
]

\addplot[
    color=blue,
    mark=square,
    ]
    coordinates {
    (0.1, 95.75757575757575)
    (0.2, 93.95973154362416)
    (0.3, 85.84905660377359)
    (0.4, 75.51867219917013)
    (0.5, 72.9064039408867)
    (0.6, 62.655601659751035)
    (0.7, 67.9245283018868)
    (0.8, 66.77852348993288)
    (0.9, 63.63636363636363)
    (1.0, 84.3501326259947)
    };
    
    \addplot[
    color=red,
    mark=square,
    ]
    coordinates {
    (0.1, 93.61702127659575)
    (0.2, 94.24460431654677)
    (0.3, 88.50758180367119)
    (0.4, 79.02654867256638)
    (0.5, 82.98022598870057)
    (0.6, 74.95575221238938)
    (0.7, 79.24980047885076)
    (0.8, 78.72559095580678)
    (0.9, 71.45390070921985)
    (1.0, 88.9839970227019)
    };
    
     \addplot[
    color=black,
    mark=square,
    ]
    coordinates {
    (0.0, 82.3751178133836)
    (0.1, 79.24454447597611)
    (0.2, 60.656659601159326)
    (0.3, 48.205646233186236)
    (0.4, 41.889855636027804)
    (0.5, 39.02091020910209)
    (0.6, 36.85900508003451)
    (0.7, 37.43952363230368)
    (0.8, 39.483126868859465)
    (0.9, 43.19185902208985)
    (1.0, 66.01008215085885)
    };
    
    % arXiv
    \addplot[
    color=orange,
    mark=square,
    ]
    coordinates {
        (0.0, 93.33333333333333)
        (0.1, 94.20101535114227)
        (0.2, 89.71152573772662)
        (0.3, 80.89219729020512)
        (0.4, 68.71181413741967)
        (0.5, 72.90627213346974)
        (0.6, 59.74100434954528)
        (0.7, 65.73305485638801)
        (0.8, 63.99470724445915)
        (0.9, 54.861425937679456)
        (1.0, 80.6093131674527)
    };
    
    % SIND
    \addplot[
    color=cyan,
    mark=square,
    ]
    coordinates {
    (0.1, 89.47368421052632)
    (0.2, 86.11298672068422)
    (0.3, 66.70731707317074)
    (0.4, 61.09772423025434)
    (0.5, 60.61427280939476)
    (0.6, 49.90629183400267)
    (0.7, 50.853658536585364)
    (0.8, 53.61242403781229)
    (0.9, 47.03947368421053)
    (1.0, 73.01681503461919)
    };

    % ROC
    \addplot[
    color=green,
    mark=square,
    ]
    coordinates {
    (0.2, 97.55501222493888)
    (0.4, 90.57660961695191)
    (0.6, 86.14506927465364)
    (0.8, 87.06193969030154)
    (1.0, 92.43072534637326)
    };
    
    %Wiki Movie Plots
    \addplot[
    color=brown,
    mark=square,
    ]
    coordinates {
    (0.1, 81.62171507607192)
    (0.2, 58.06315789473684)
    (0.3, 43.685061186792886)
    (0.4, 40.25632599408478)
    (0.5, 32.94582145207282)
    (0.6, 27.634817632579995)
    (0.7, 26.668205957053797)
    (0.8, 27.396373056994815)
    (0.9, 30.082606700321247)
    (1.0, 44.614003590664275)
    };
    \legend{NeurIPS, AAN, NSF, arXiv, SIND, ROC, Movie Plots}
\end{axis}
\label{position-wise}
\end{tikzpicture}

%% file: figures/acc_first_last.tex
		\begin{tabular}{ l c c |c c} % <-- Alignments: 1st column left, 2nd middle and 3rd right, with vertical lines in between
			\toprule[1pt]
			Method & \multicolumn{2}{c}{\textbf{arXiv}} & \multicolumn{2}{c}{\textbf{SIND}} \TBstrut\\
			\cline{2-5}
			& {head} & {tail} & {head} & {tail} \TBstrut\\
			\midrule[1pt]
			Pairwise Model & 84.85 & 62.37 & - & - \Tstrut\\
            LSTM+PtrNet & 90.47 & 66.49 & 74.66 & 53.30 \Tstrut\\
            ATTOrderNet & 91.00 & 68.08 & 76.00 & 54.42 \Tstrut\\
            SE-Graph & 92.28 & 70.45 & 78.12 & 56.68 \Tstrut\\
            FUDecoder & 92.76 & 71.49 & 78.08 & 57.32 \Tstrut\\
            TGCM & 92.46 & 69.45 & 78.98 & 56.24 \Tstrut\\
            RankTxNet & 92.97 & 69.13 & 80.32 & 59.68 \Tstrut\\
            BERSON & 94.75 & 76.69 & 84.95 & 64.87 \Tstrut\\
            \hline
            % \textcolor{red}{\textbf{96.67}} & \textcolor{red}{\textbf{81.67}}
            {\rebart} & \textbf{96.46} & \textbf{80.62} & \textbf{87.97} & \textbf{73.02} \TBstrut\\
			\bottomrule[1pt]
		\end{tabular}

%% file: figures/degree_correctly_predicted.tex
    \begin{tikzpicture}[scale=0.5]
\begin{axis}[
	ytick={0, 20, 40, 60, 80, 100, 120, 140, 160, 180},
	xlabel = {\Large $d(Y, Y^*)$},
	ylabel= {\Large \# of instances},
	enlargelimits=0.05,
% 	ybar interval=0.7,
	ymin=0, ymax=180,
	xmin=1, xmax=18,
    xtick={2,4,6,8,10,12, 14, 16, 18},
    grid style=dashed,
    ymajorgrids=true,
]
\addplot[
        color=brown,
        %fill=green!25,
        mark=square,
    ]
	coordinates {(1, 152)
                    (2, 170)
                    (3, 112)
                    (4, 103)
                    (5, 67)
                    (6, 81)
                    (7, 64)
                    (8, 56)
                    (9, 46)
                    (10, 38)
                    (11, 31)
                    (12, 16)
                    (13, 13)
                    (14, 9)
                    (15, 4)
                    (16, 4)
                    (17, 0)
                    (18, 0)
                    (19, 0)};
\end{axis}
\end{tikzpicture}

%% file: figures/acc_wrt_distance_from_orig_position.tex
\begin{tikzpicture}[scale=0.55]
\begin{axis}[
    xlabel={\Large $\delta^{rel}(s_i)$},
    ylabel={\Large Accuracy},
    xmin=0, xmax=0.9,
    ymin=0, ymax=100,
    xtick={0, 0.1, 0.2, 0.3, 0.4, 0.5, 0.6, 0.7, 0.8, 0.9},
    ytick={0, 20, 30, 40, 50, 60, 70, 80, 90, 100},
    ymajorgrids=true,
    grid style=dashed,
    legend pos=outer north east
]

% NIPS
\addplot[
    color=blue,
    mark=square,
    ]
    coordinates {
    (0.0, 82.21649484536083)
    (0.1, 71.33105802047781)
    (0.2, 74.13441955193483)
    (0.3, 78.21428571428571)
    (0.4, 70.50359712230215)
    (0.5, 78.81773399014779)
    (0.6, 81.15183246073299)
    (0.7, 83.72093023255815)
    (0.8, 88.98305084745762)
    (0.9, 78.04878048780488)
    };
    
    % AAN
    \addplot[
    color=red,
    mark=square,
    ]
    coordinates {
    (0.0, 87.2434017595308)
    (0.1, 74.20675537359263)
    (0.2, 83.51046698872786)
    (0.3, 83.77906976744185)
    (0.4, 79.58872810357958)
    (0.5, 85.55798687089715)
    (0.6, 81.85654008438819)
    (0.7, 88.98623279098874)
    (0.8, 93.05555555555556)
    (0.9, 91.17647058823529)
    };
    
    % NSF
     \addplot[
    color=black,
    mark=square,
    ]
    coordinates {
    (0.0, 55.20032194598462)
    (0.1, 43.79858705436295)
    (0.2, 46.944610859974944)
    (0.3, 46.44626195138809)
    (0.4, 46.828539107950874)
    (0.5, 48.949694592044494)
    (0.6, 52.253429131286744)
    (0.7, 55.97113438352548)
    (0.8, 63.035659233377864)
    (0.9, 70.21825862473598)
% (1.0, 54.90196078431373)
    };
    
     % Arxiv 
     \addplot[
    color=orange,
    mark=square,
    ]
    coordinates {
    (0.0, 79.52759465740665)
    (0.1, 61.1113974061668)
    (0.2, 72.19859109937961)
    (0.3, 73.40401682689136)
    (0.4, 68.56088194398464)
    (0.5, 77.25686019754865)
    (0.6, 72.86956521739131)
    (0.7, 80.7653743315508)
    (0.8, 83.82965495803543)
    (0.9, 82.67637459201606)
    % (1.0, 100.0)
    };
    
    % SIND
    \addplot[
    color=cyan,
    mark=square,
    ]
    coordinates {
    (0.0, 66.64079177178343)
    (0.1, 48.669201520912544)
    (0.2, 61.50568181818182)
    (0.3, 61.165919282511204)
    (0.4, 61.4376130198915)
    (0.5, 70.63063063063063)
    (0.6, 67.4164737016209)
    (0.7, 74.50980392156863)
    (0.8, 79.08020190689848)
    (0.9, 76.82926829268293)
    };

    % ROC
    \addplot[
    color=green,
    mark=square,
    ]
    coordinates {
    (0.0, 91.2157268494568)
    (0.2, 89.80907826526742)
    (0.4, 89.76161893328761)
    (0.6, 91.67296548248929)
    (0.8, 94.44583437578183)
    };
    
    %Wiki Movie Plots
    \addplot[
    color=brown,
    mark=square,
    ]
    coordinates {
        (0.0, 53.7515006002401)
    (0.1, 35.91387559808613)
    (0.2, 40.171858216971)
    (0.3, 38.60802157061005)
    (0.4, 42.63070077864294)
    (0.5, 41.69007996542036)
    (0.6, 43.88663967611336)
    (0.7, 45.415005436752445)
    (0.8, 53.278688524590166)
    (0.9, 61.50684931506849)
    };
    \legend{NeurIPS, AAN, NSF, arXiv, SIND, ROC, Movie Plots}
\end{axis}
\label{dist-wise}
\end{tikzpicture}

%% file: emnlp2020.bbl
\begin{thebibliography}{35}
\expandafter\ifx\csname natexlab\endcsname\relax\def\natexlab#1{#1}\fi

\bibitem[{Barzilay and Elhadad(2002)}]{barzilay2002inferring}
Regina Barzilay and Noemie Elhadad. 2002.
\newblock Inferring strategies for sentence ordering in multidocument news
  summarization.
\newblock \emph{Journal of Artificial Intelligence Research}, 17:35--55.

\bibitem[{Barzilay and Lapata(2005)}]{barzilay2008modeling}
Regina Barzilay and Mirella Lapata. 2005.
\newblock \href {https://doi.org/10.3115/1219840.1219858} {Modeling local
  coherence: An entity-based approach}.
\newblock In \emph{Proceedings of the 43rd Annual Meeting of the Association
  for Computational Linguistics ({ACL}{'}05)}, pages 141--148, Ann Arbor,
  Michigan. Association for Computational Linguistics.

\bibitem[{Chen et~al.(2016)Chen, Qiu, and Huang}]{chen2016neural}
Xinchi Chen, Xipeng Qiu, and Xuanjing Huang. 2016.
\newblock Neural sentence ordering.
\newblock \emph{arXiv preprint arXiv:1607.06952}.

\bibitem[{Cui et~al.(2018)Cui, Li, Chen, and Zhang}]{cui2018deep}
Baiyun Cui, Yingming Li, Ming Chen, and Zhongfei Zhang. 2018.
\newblock \href {https://doi.org/10.18653/v1/D18-1465} {Deep attentive sentence
  ordering network}.
\newblock In \emph{Proceedings of the 2018 Conference on Empirical Methods in
  Natural Language Processing}, pages 4340--4349, Brussels, Belgium.
  Association for Computational Linguistics.

\bibitem[{Cui et~al.(2020)Cui, Li, and Zhang}]{cui2020bert}
Baiyun Cui, Yingming Li, and Zhongfei Zhang. 2020.
\newblock \href {https://doi.org/10.18653/v1/2020.emnlp-main.511}
  {{BERT}-enhanced relational sentence ordering network}.
\newblock In \emph{Proceedings of the 2020 Conference on Empirical Methods in
  Natural Language Processing (EMNLP)}, pages 6310--6320, Online. Association
  for Computational Linguistics.

\bibitem[{Devlin et~al.(2019)Devlin, Chang, Lee, and
  Toutanova}]{devlin2018bert}
Jacob Devlin, Ming-Wei Chang, Kenton Lee, and Kristina Toutanova. 2019.
\newblock \href {https://doi.org/10.18653/v1/N19-1423} {{BERT}: Pre-training of
  deep bidirectional transformers for language understanding}.
\newblock In \emph{Proceedings of the 2019 Conference of the North {A}merican
  Chapter of the Association for Computational Linguistics: Human Language
  Technologies, Volume 1 (Long and Short Papers)}, pages 4171--4186,
  Minneapolis, Minnesota. Association for Computational Linguistics.

\bibitem[{Elsner and Charniak(2008)}]{elsner-charniak-2008-coreference}
Micha Elsner and Eugene Charniak. 2008.
\newblock \href {https://www.aclweb.org/anthology/P08-2011}
  {Coreference-inspired coherence modeling}.
\newblock In \emph{Proceedings of ACL-08: HLT, Short Papers}, pages 41--44,
  Columbus, Ohio. Association for Computational Linguistics.

\bibitem[{Elsner and Charniak(2011)}]{elsner2011extending}
Micha Elsner and Eugene Charniak. 2011.
\newblock Extending the entity grid with entity-specific features.
\newblock In \emph{Proceedings of the 49th Annual Meeting of the Association
  for Computational Linguistics: Human Language Technologies}, pages 125--129.

\bibitem[{Galanis et~al.(2012)Galanis, Lampouras, and
  Androutsopoulos}]{galanis2012extractive}
Dimitrios Galanis, Gerasimos Lampouras, and Ion Androutsopoulos. 2012.
\newblock \href {https://www.aclweb.org/anthology/C12-1056} {Extractive
  multi-document summarization with integer linear programming and support
  vector regression}.
\newblock In \emph{Proceedings of {COLING} 2012}, pages 911--926, Mumbai,
  India. The COLING 2012 Organizing Committee.

\bibitem[{Gong et~al.(2016)Gong, Chen, Qiu, and Huang}]{gong2016end}
Jingjing Gong, Xinchi Chen, Xipeng Qiu, and Xuanjing Huang. 2016.
\newblock End-to-end neural sentence ordering using pointer network.
\newblock \emph{arXiv preprint arXiv:1611.04953}.

\bibitem[{Guinaudeau and Strube(2013)}]{guinaudeau2013graph}
Camille Guinaudeau and Michael Strube. 2013.
\newblock Graph-based local coherence modeling.
\newblock In \emph{Proceedings of the 51st Annual Meeting of the Association
  for Computational Linguistics (Volume 1: Long Papers)}, pages 93--103.

\bibitem[{Holtzman et~al.(2020)Holtzman, Buys, Du, Forbes, and
  Choi}]{holtzman2019curious}
Ari Holtzman, Jan Buys, Li~Du, Maxwell Forbes, and Yejin Choi. 2020.
\newblock \href {https://openreview.net/forum?id=rygGQyrFvH} {The curious case
  of neural text degeneration}.
\newblock In \emph{8th International Conference on Learning Representations,
  {ICLR} 2020, Addis Ababa, Ethiopia, April 26-30, 2020}. OpenReview.net.

\bibitem[{Huang et~al.(2016)Huang, Ferraro, Mostafazadeh, Misra, Agrawal,
  Devlin, Girshick, He, Kohli, Batra, Zitnick, Parikh, Vanderwende, Galley, and
  Mitchell}]{huang2016visual}
Ting-Hao~Kenneth Huang, Francis Ferraro, Nasrin Mostafazadeh, Ishan Misra,
  Aishwarya Agrawal, Jacob Devlin, Ross Girshick, Xiaodong He, Pushmeet Kohli,
  Dhruv Batra, C.~Lawrence Zitnick, Devi Parikh, Lucy Vanderwende, Michel
  Galley, and Margaret Mitchell. 2016.
\newblock \href {https://doi.org/10.18653/v1/N16-1147} {Visual storytelling}.
\newblock In \emph{Proceedings of the 2016 Conference of the North {A}merican
  Chapter of the Association for Computational Linguistics: Human Language
  Technologies}, pages 1233--1239, San Diego, California. Association for
  Computational Linguistics.

\bibitem[{Kobayashi et~al.(2020)Kobayashi, Kuribayashi, Yokoi, and
  Inui}]{kobayashi2020attention}
Goro Kobayashi, Tatsuki Kuribayashi, Sho Yokoi, and Kentaro Inui. 2020.
\newblock \href {https://doi.org/10.18653/v1/2020.emnlp-main.574} {Attention is
  not only a weight: Analyzing transformers with vector norms}.
\newblock In \emph{Proceedings of the 2020 Conference on Empirical Methods in
  Natural Language Processing (EMNLP)}, pages 7057--7075, Online. Association
  for Computational Linguistics.

\bibitem[{Kumar et~al.(2020)Kumar, Brahma, Karnick, and Rai}]{kumar2020deep}
Pawan Kumar, Dhanajit Brahma, Harish Karnick, and Piyush Rai. 2020.
\newblock \href {https://aaai.org/ojs/index.php/AAAI/article/view/6323} {Deep
  attentive ranking networks for learning to order sentences}.
\newblock In \emph{The Thirty-Fourth {AAAI} Conference on Artificial
  Intelligence, {AAAI} 2020, The Thirty-Second Innovative Applications of
  Artificial Intelligence Conference, {IAAI} 2020, The Tenth {AAAI} Symposium
  on Educational Advances in Artificial Intelligence, {EAAI} 2020, New York,
  NY, USA, February 7-12, 2020}, pages 8115--8122. {AAAI} Press.

\bibitem[{Lapata and Barzilay(2005)}]{LapataB05}
Mirella Lapata and Regina Barzilay. 2005.
\newblock \href {http://ijcai.org/Proceedings/05/Papers/0505.pdf} {Automatic
  evaluation of text coherence: Models and representations}.
\newblock In \emph{IJCAI-05, Proceedings of the Nineteenth International Joint
  Conference on Artificial Intelligence, Edinburgh, Scotland, UK, July 30 -
  August 5, 2005}, pages 1085--1090. Professional Book Center.

\bibitem[{Lewis et~al.(2020)Lewis, Liu, Goyal, Ghazvininejad, Mohamed, Levy,
  Stoyanov, and Zettlemoyer}]{lewis2019bart}
Mike Lewis, Yinhan Liu, Naman Goyal, Marjan Ghazvininejad, Abdelrahman Mohamed,
  Omer Levy, Veselin Stoyanov, and Luke Zettlemoyer. 2020.
\newblock \href {https://doi.org/10.18653/v1/2020.acl-main.703} {{BART}:
  Denoising sequence-to-sequence pre-training for natural language generation,
  translation, and comprehension}.
\newblock In \emph{Proceedings of the 58th Annual Meeting of the Association
  for Computational Linguistics}, pages 7871--7880, Online. Association for
  Computational Linguistics.

\bibitem[{Li and Hovy(2014)}]{li-hovy-2014-model}
Jiwei Li and Eduard Hovy. 2014.
\newblock \href {https://doi.org/10.3115/v1/D14-1218} {A model of coherence
  based on distributed sentence representation}.
\newblock In \emph{Proceedings of the 2014 Conference on Empirical Methods in
  Natural Language Processing ({EMNLP})}, pages 2039--2048, Doha, Qatar.
  Association for Computational Linguistics.

\bibitem[{Liu et~al.(2018)Liu, Shen, Duh, and Gao}]{liu2017stochastic}
Xiaodong Liu, Yelong Shen, Kevin Duh, and Jianfeng Gao. 2018.
\newblock \href {https://doi.org/10.18653/v1/P18-1157} {Stochastic answer
  networks for machine reading comprehension}.
\newblock In \emph{Proceedings of the 56th Annual Meeting of the Association
  for Computational Linguistics (Volume 1: Long Papers)}, pages 1694--1704,
  Melbourne, Australia. Association for Computational Linguistics.

\bibitem[{Logeswaran et~al.(2018{\natexlab{a}})Logeswaran, Lee, and
  Radev}]{logeswaran2018sentence}
Lajanugen Logeswaran, Honglak Lee, and Dragomir~R. Radev. 2018{\natexlab{a}}.
\newblock \href
  {https://www.aaai.org/ocs/index.php/AAAI/AAAI18/paper/view/17011} {Sentence
  ordering and coherence modeling using recurrent neural networks}.
\newblock In \emph{Proceedings of the Thirty-Second {AAAI} Conference on
  Artificial Intelligence, (AAAI-18), the 30th innovative Applications of
  Artificial Intelligence (IAAI-18), and the 8th {AAAI} Symposium on
  Educational Advances in Artificial Intelligence (EAAI-18), New Orleans,
  Louisiana, USA, February 2-7, 2018}, pages 5285--5292. {AAAI} Press.

\bibitem[{Logeswaran et~al.(2018{\natexlab{b}})Logeswaran, Lee, and
  Radev}]{LogeswaranLR18}
Lajanugen Logeswaran, Honglak Lee, and Dragomir~R. Radev. 2018{\natexlab{b}}.
\newblock \href
  {https://www.aaai.org/ocs/index.php/AAAI/AAAI18/paper/view/17011} {Sentence
  ordering and coherence modeling using recurrent neural networks}.
\newblock In \emph{Proceedings of the Thirty-Second {AAAI} Conference on
  Artificial Intelligence, (AAAI-18), the 30th innovative Applications of
  Artificial Intelligence (IAAI-18), and the 8th {AAAI} Symposium on
  Educational Advances in Artificial Intelligence (EAAI-18), New Orleans,
  Louisiana, USA, February 2-7, 2018}, pages 5285--5292. {AAAI} Press.

\bibitem[{Louis and Nenkova(2012)}]{louis-nenkova-2012-coherence}
Annie Louis and Ani Nenkova. 2012.
\newblock \href {https://www.aclweb.org/anthology/D12-1106} {A coherence model
  based on syntactic patterns}.
\newblock In \emph{Proceedings of the 2012 Joint Conference on Empirical
  Methods in Natural Language Processing and Computational Natural Language
  Learning}, pages 1157--1168, Jeju Island, Korea. Association for
  Computational Linguistics.

\bibitem[{McInnes et~al.(2018)McInnes, Healy, and Melville}]{mcinnes2018umap}
Leland McInnes, John Healy, and James Melville. 2018.
\newblock Umap: Uniform manifold approximation and projection for dimension
  reduction.
\newblock \emph{arXiv preprint arXiv:1802.03426}.

\bibitem[{Nallapati et~al.(2017)Nallapati, Zhai, and
  Zhou}]{nallapati2017summarunner}
Ramesh Nallapati, Feifei Zhai, and Bowen Zhou. 2017.
\newblock \href {http://aaai.org/ocs/index.php/AAAI/AAAI17/paper/view/14636}
  {Summarunner: {A} recurrent neural network based sequence model for
  extractive summarization of documents}.
\newblock In \emph{Proceedings of the Thirty-First {AAAI} Conference on
  Artificial Intelligence, February 4-9, 2017, San Francisco, California,
  {USA}}, pages 3075--3081. {AAAI} Press.

\bibitem[{Oh et~al.(2019)Oh, Seo, Shin, Jo, and Lee}]{oh2019topic}
Byungkook Oh, Seungmin Seo, Cheolheon Shin, Eunju Jo, and Kyong-Ho Lee. 2019.
\newblock \href {https://doi.org/10.18653/v1/D19-1232} {Topic-guided coherence
  modeling for sentence ordering by preserving global and local information}.
\newblock In \emph{Proceedings of the 2019 Conference on Empirical Methods in
  Natural Language Processing and the 9th International Joint Conference on
  Natural Language Processing (EMNLP-IJCNLP)}, pages 2273--2283, Hong Kong,
  China. Association for Computational Linguistics.

\bibitem[{Prabhumoye et~al.(2020)Prabhumoye, Salakhutdinov, and
  Black}]{prabhumoye2020topological}
Shrimai Prabhumoye, Ruslan Salakhutdinov, and Alan~W Black. 2020.
\newblock \href {https://doi.org/10.18653/v1/2020.acl-main.248} {Topological
  sort for sentence ordering}.
\newblock In \emph{Proceedings of the 58th Annual Meeting of the Association
  for Computational Linguistics}, pages 2783--2792, Online. Association for
  Computational Linguistics.

\bibitem[{Ravfogel et~al.(2020)Ravfogel, Elazar, Goldberger, and
  Goldberg}]{ravfogel2020unsupervised}
Shauli Ravfogel, Yanai Elazar, Jacob Goldberger, and Yoav Goldberg. 2020.
\newblock Unsupervised distillation of syntactic information from
  contextualized word representations.
\newblock \emph{arXiv preprint arXiv:2010.05265}.

\bibitem[{Schwartz et~al.(2017)Schwartz, Sap, Konstas, Zilles, Choi, and
  Smith}]{schwartz2017effect}
Roy Schwartz, Maarten Sap, Ioannis Konstas, Leila Zilles, Yejin Choi, and
  Noah~A. Smith. 2017.
\newblock \href {https://doi.org/10.18653/v1/K17-1004} {The effect of different
  writing tasks on linguistic style: A case study of the {ROC} story cloze
  task}.
\newblock In \emph{Proceedings of the 21st Conference on Computational Natural
  Language Learning ({C}o{NLL} 2017)}, pages 15--25, Vancouver, Canada.
  Association for Computational Linguistics.

\bibitem[{Vinyals et~al.(2015)Vinyals, Fortunato, and
  Jaitly}]{vinyals2015pointer}
Oriol Vinyals, Meire Fortunato, and Navdeep Jaitly. 2015.
\newblock \href
  {https://proceedings.neurips.cc/paper/2015/hash/29921001f2f04bd3baee84a12e98098f-Abstract.html}
  {Pointer networks}.
\newblock In \emph{Advances in Neural Information Processing Systems 28: Annual
  Conference on Neural Information Processing Systems 2015, December 7-12,
  2015, Montreal, Quebec, Canada}, pages 2692--2700.

\bibitem[{Wang and Wan(2019)}]{wang2019hierarchical}
Tianming Wang and Xiaojun Wan. 2019.
\newblock \href {https://doi.org/10.1609/aaai.v33i01.33017184} {Hierarchical
  attention networks for sentence ordering}.
\newblock In \emph{Proceedings of the AAAI Conference on Artificial
  Intelligence}, 01, pages 7184--7191. {AAAI} Press.

\bibitem[{Wolf et~al.(2019)Wolf, Debut, Sanh, Chaumond, Delangue, Moi, Cistac,
  Rault, Louf, Funtowicz, and Brew}]{Wolf2019HuggingFacesTS}
Thomas Wolf, Lysandre Debut, Victor Sanh, Julien Chaumond, Clement Delangue,
  Anthony Moi, Pierric Cistac, Tim Rault, R'emi Louf, Morgan Funtowicz, and
  Jamie Brew. 2019.
\newblock Huggingface's transformers: State-of-the-art natural language
  processing.
\newblock \emph{arXiv}.

\bibitem[{Yin et~al.(2020)Yin, Meng, Su, Ge, Song, Zhou, and
  Luo}]{yin2020enhancing}
Yongjing Yin, Fandong Meng, Jinsong Su, Yubin Ge, Lingeng Song, Jie Zhou, and
  Jiebo Luo. 2020.
\newblock Enhancing pointer network for sentence ordering with pairwise
  ordering predictions.
\newblock In \emph{Proceedings of the AAAI Conference on Artificial
  Intelligence}, volume~34, pages 9482--9489. {AAAI} Press.

\bibitem[{Yin et~al.(2019)Yin, Song, Su, Zeng, Zhou, and Luo}]{yin2019graph}
Yongjing Yin, Linfeng Song, Jinsong Su, Jiali Zeng, Chulun Zhou, and Jiebo Luo.
  2019.
\newblock \href {https://doi.org/10.24963/ijcai.2019/748} {Graph-based neural
  sentence ordering}.
\newblock In \emph{Proceedings of the Twenty-Eighth International Joint
  Conference on Artificial Intelligence, {IJCAI} 2019, Macao, China, August
  10-16, 2019}, pages 5387--5393. ijcai.org.

\bibitem[{Yu et~al.(2018)Yu, Dohan, Luong, Zhao, Chen, Norouzi, and
  Le}]{yu2018qanet}
Adams~Wei Yu, David Dohan, Minh{-}Thang Luong, Rui Zhao, Kai Chen, Mohammad
  Norouzi, and Quoc~V. Le. 2018.
\newblock \href {https://openreview.net/forum?id=B14TlG-RW} {Qanet: Combining
  local convolution with global self-attention for reading comprehension}.
\newblock In \emph{6th International Conference on Learning Representations,
  {ICLR} 2018, Vancouver, BC, Canada, April 30 - May 3, 2018, Conference Track
  Proceedings}. OpenReview.net.

\bibitem[{Zhu et~al.(2021)Zhu, Zhou, Nie, Liu, and Dou}]{zhu2021neural}
Yutao Zhu, Kun Zhou, Jian-Yun Nie, Shengchao Liu, and Zhicheng Dou. 2021.
\newblock Neural sentence ordering based on constraint graphs.
\newblock In \emph{Proceedings of the AAAI Conference on Artificial
  Intelligence}, volume~35, pages 14656--14664.

\end{thebibliography}
